\documentclass[journal]{IEEEtran}

\usepackage{amsmath}
\usepackage{amssymb}
\usepackage{amsfonts}
\usepackage{graphicx}
\usepackage{epsfig}
\usepackage{subfigure}
\usepackage{psfrag}
\usepackage{cite}
\usepackage{latexsym}
\usepackage{url}
\usepackage{color}
\usepackage{multirow}
\usepackage{mathtools}
\usepackage{bm}
\usepackage{booktabs}
\usepackage{algorithm}
\usepackage{algpseudocode}
\usepackage{bm}

\usepackage{verbatim}
\usepackage{indentfirst}
\usepackage{hyperref}
\usepackage{multirow}
\usepackage{bbm}

\graphicspath{{fig/}}

\PassOptionsToPackage{bookmarks={false}}{hyperref}

\IEEEoverridecommandlockouts

\newtheorem{theorem}{\underline{Theorem}}[section]
\newtheorem{lemma}{\underline{Lemma}}[section]

\newtheorem{proposition}{\underline{Proposition}}[section]

\newtheorem{remark}{\underline{Remark}}[section]
\newcommand{\mv}[1]{\mbox{\boldmath{$ #1 $}}}

\allowdisplaybreaks[4] 

\begin{document}
\title{The Larger the Merrier? Efficient Large AI Model Inference in Wireless Edge Networks}
\author{
Zhonghao Lyu, Ming Xiao, \emph{Senior Member, IEEE,} Jie Xu, \emph{Fellow, IEEE,} \\ Mikael Skoglund, \emph{Fellow, IEEE,} and Marco Di Renzo, \emph{Fellow, IEEE}\\
\thanks{Z. Lyu, M. Xiao, and M. Skoglund are with the Department of Information Science and Engineering, KTH Royal Institute of Technology, Stockholm, Sweden (e-mail: lzhon@kth.se, mingx@kth.se, skoglund@kth.se).}
\thanks{J. Xu is with the School of Science and Engineering, the Shenzhen Future Network of Intelligence Institute, and the Guangdong Provincial Key Laboratory of Future Networks of Intelligence, The Chinese University of Hong Kong (Shenzhen), Guangdong 518172, China  (e-mail: xujie@cuhk.edu.cn). \emph{(Corresponding author: M. Xiao and J. Xu.)}}
\thanks{M. Di Renzo is with Universit\'e Paris-Saclay, CNRS, CentraleSup\'elec, Laboratoire des Signaux et Syst\`emes, 3 Rue Joliot-Curie, 91192 Gif-sur-Yvette, France. (marco.di-renzo@universite-paris-saclay.fr), and with King's College London, Centre for Telecommunications Research -- Department of Engineering, WC2R 2LS London, United Kingdom (marco.di\_renzo@kcl.ac.uk).}
}

\maketitle

\begin{abstract}
The growing demand for large artificial intelligence model (LAIM) services is driving a paradigm shift from traditional cloud-based inference to edge-based inference for low-latency, privacy-preserving applications. In particular, edge-device co-inference, which partitions LAIMs between edge devices and servers, has emerged as a promising strategy for resource-efficient LAIM execution in wireless networks.
In this paper, we investigate a pruning-aware LAIM co-inference scheme, where a pre-trained LAIM is pruned and partitioned into on-device and on-server sub-models for deployment. For analysis, we first prove that the LAIM output distortion is upper bounded by its parameter distortion. Then, we derive a lower bound on parameter distortion via rate-distortion theory, analytically capturing the relationship between pruning ratio and co-inference performance. 
Next, based on the analytical results, we formulate an LAIM co-inference distortion bound minimization problem by jointly optimizing the pruning ratio, transmit power, and computation frequency under system latency, energy, and available resource constraints. Moreover, we propose an efficient algorithm to tackle the considered highly non-convex problem. 
Finally, extensive simulations demonstrate the effectiveness of the proposed design. In particular, model parameter distortion is shown to provide a reliable bound on output distortion. Also, the proposed joint pruning ratio and resource management design achieves superior performance in balancing trade-offs among inference performance, system latency, and energy consumption compared with benchmark schemes, such as fully on-device and on-server inference. Moreover, the split point is shown to play a critical role in system performance optimization under heterogeneous and resource-limited edge environments.
\end{abstract}

\section{Introduction}
\subsection{Backgrounds and Related Works}
The rapid advancement of large artificial intelligence models (LAIMs) has marked a significant breakthrough in AI, drawing increasing attention from academia, industry, and the public. Built upon the transformer architecture, LAIMs leverage massive-scale parameters and extensive training data to achieve remarkable ability in  understanding, reasoning, and generating human-like content \cite{LBariah2024, LDong2025}. Representative examples, such as OpenAI's GPT-4 \cite{JAchiam2023}, have transcended conventional AI limitations to tackle complex tasks across diverse domains, such as multi-modal data processing. Their widespread adoption has led to transformative applications, ranging from AI-powered virtual assistants to autonomous control systems. 

However, the massive model size and high computational complexity of LAIMs pose significant challenges for efficient inference, particularly for latency-sensitive services in resource-limited scenarios \cite{YDong2024}. Currently, most LAIM inference is conducted through centralized cloud-based paradigms (i.e., on-server inference), where user data is transmitted to remote cloud servers for processing. Although efforts such as graphic processing unit (GPU) scheduling optimization have been proposed to accelerate cloud-based LAIM inference \cite{TGriggs2024},  this paradigm still suffers from inherent limitations, such as high communication latency and potential privacy risks.

 
Alternatively, with the surging of edge computing for future mobile networks, e.g., the sixth-generation (6G) networks, the computation capability of edge servers and devices has continuously grown \cite{MChen2021}, making edge-based LAIM deployments increasingly feasible \cite{HZhou,KBLetaief}. A few recent works have investigated on-device inference. For example, EdgeMoE \cite{RYi2023} and LiteMoE \cite{YZhuang2024} are memory- and computation-efficient solutions for deploying LAIMs directly on edge devices for mobile applications. However, on-device inference is severely constrained by the limited memory, computation, and battery capacity of edge devices. For example, an inference task on GPT 6.7B (with FP16 precision, batch size 64, and input/output sequence length 512/32) requires 41.84 GB of running memory \cite{Nvidia2023}, far beyond what most existing edge devices can support.

To address these limitations, collaborative inference (co-inference), also known as split inference, has emerged as a promising solution, where LAIMs are partitioned into sub-models executed across edge devices and edge/cloud servers \cite{GZhu2023}. Early works on co-inference have mainly focused on regular-scale AI models, exploring two primary strategies: task-oriented intermediate feature encoding and  joint task offloading and resource allocation. For {\it task-oriented intermediate feature encoding}, raw data is encoded into compressed semantic features to reduce communication overhead. For instance, leveraging deep neural networks (DNNs) as powerful encoders has inspired the field of semantic/task-oriented communications \cite{ZQin2024}, enabling various applications such as data reconstruction  \cite{JDai2022, HDu2023}, intelligent task execution \cite{Shao2023, HZhou2024}, and multi-task services \cite{ZLyu2024}. For {\it joint task-offloading and resource allocation}, AI inference workloads are partitioned and allocated across distributed processing ends based on available communication and computation resources \cite{WRen2023}. For example, \cite{ELi2020,JYan2022,WShi2019} have considered model splitting and placement optimization in edge-device co-inference systems. Moreover, the authors of \cite{ZLiu2023,SFYilmaz} have further studied multi-user co-inference systems with joint communication and computation resource management.

However, directly applying the mentioned strategies to LAIMs is highly challenging. LAIM co-inference should consider their unique challenges of massive parameter scales and computation intensity. To this end, recent efforts have considered {\it large-scale joint task-offloading and resource management}, and {\it model compression} for efficient LAIM co-inference.  For instance, \cite{YHe2024} has explored cloud-edge active co-inference for LAIMs but underutilized the abundant computation potential of edge devices. To exploit this potential,  \cite{YChen2024,SOh2024,Mzhang} have studied LAIM co-inference between edge devices and servers. Specifically, \cite{YChen2024} has analyzed the impact of different model splitting points under dynamic wireless channels. Also, \cite{SOh2024} has introduced a hybrid architecture involving a small model on the device and a full LAIM on the server, with local output quality-aware collaboration. Different from these single-device settings, \cite{Mzhang} has extended to a general multi-device multi-server co-inference framework with joint device selection and model partitioning.

Complementary to large-scale joint task-offloading and resource management, model compression is another crucial technique for reducing the computation and memory burden of LAIM inference. Key methods include model pruning \cite{Bisik, AHGadhikar,JLee}, quantization \cite{GXiao}, and knowledge distillation  \cite{YGu}. Model pruning removes redundant parameters based on random \cite{AHGadhikar} or magnitude-based \cite{JLee} criteria, quantization reduces numerical precision using low-bit representations \cite{GXiao}, and knowledge distillation trains compact student models to replicate full LAIM outputs \cite{YGu}. These techniques can be integrated within LAIM co-inference systems to further enhance feasibility and performance at resource-limited edge environments.

\subsection{Motivations and Contributions}

Despite the progress, most existing works on LAIM inference adopt empirical or heuristic approaches to model compression for fast inference, lacking theoretical foundations on inference performance or resource efficiency. A fundamental research challenge remains unresolved: To what extent should LAIMs be compressed to well balance model complexity and inference quality? The challenge becomes particularly critical in resource-limited edge inference systems, where improper compression can severely affect system performance. Specifically, overly aggressive compression may significantly degrade inference quality, while under-compression may make the system infeasible, violating quality-of-service (QoS) requirements. However, theoretically characterizing the relationship between the compression ratio and inference performance is highly non-trivial. LAIMs are composed of complex, deeply stacked non-linear layers, which make the analysis of model compression induced distortion analytically elusive.

Furthermore, under practical QoS requirements, there generally exist inherent trade-offs among inference performance, system delay, and energy consumption. To effectively balance these trade-offs, it is essential to jointly optimize the compression strategies together with the communication and computation resource allocation. However, for LAIM edge co-inference, different model segments are executed across heterogeneous devices and servers with asymmetric communication and computation capabilities.
Such heterogeneity makes the LAIM co-inference design even more challenging, which have not been investigated in the existing literature yet.

To tackle the above challenges, in this work, we investigate a pruning-aware LAIM co-inference system from the theoretical and design aspects. Specifically, the pre-trained LAIM is partitioned into two sub-models deployed on the edge device and server, respectively. The device performs partial inference to extract intermediate embeddings, which are then transmitted to the edge server for final inference. To alleviate the computation and storage burden, we adopt model compression, particularly  model pruning, to reduce the complexity of both the on-device and on-server models prior to deployment. We then theoretically characterize the relationship between the pruning ratio and inference distortion, which provides a useful guideline for the LAIM co-inference system design. The main contributions are summarized as follows.
\begin{itemize}
\item To address the analytical intractability of LAIM output distortion by model pruning, we first prove that the inference output distortion is upper bounded by the model parameter distortion, starting with fully-connected (FC) DNNs and extending to general AI models. We then derive a lower bound on inference quality distortion based on rate-distortion theory, which captures the influence of various important system parameters (such as pruning ratio and model entropy) on inference distortion. 
\item Building upon the theoretical distortion bound, we formulate an inference distortion bound minimization problem by jointly optimizing the pruning ratio, transmit power, and computation frequency, subject to various constraints on the system resources (i.e., the maximum transmission power and computation clock frequency), latency, and energy consumption. However, the considered problem is highly non-convex and hard to solve due to the close coupling between the model pruning ratio and the computation/communication variables. To address the problem, we develop an efficient algorithm to obtain a high-quality sub-optimal solution.
\item Extensive simulations demonstrate the superiority of the proposed joint model pruning and communication-computation resource allocation design. Firstly, we show that LAIM parameter distortion is a reliable performance upper bound for the output distortion. Secondly, it is shown that the proposed design achieves improved inference quality compared with various benchmark schemes, effectively balancing the trade-offs between inference performance, system latency, and energy consumption via better leveraging system resources with adaptive pruning strategies. Finally, we show that the proposed LAIM co-inference design outperforms conventional paradigms such as fully on-device and on-server inference, and the split point plays an important role in the optimized system performance under heterogeneous and resource-limited edge environments.
\end{itemize}

The remainder of this paper is organized as follows. Section II introduces the system model for LAIM co-inference. Section III presents the approximation of LAIM output distortion. Section IV provides the rate-distortion analysis for model pruning. Section V formulates the joint pruning ratio, communication, and computation design problem, and develops an efficient algorithm to solve it. Section VI presents numerical results to validate the effectiveness of the proposed LAIM co-inference design. Section VII concludes the paper.

\begin{figure}[h]
	\centering
	 \epsfxsize=1\linewidth
		\includegraphics[width=8.5cm]{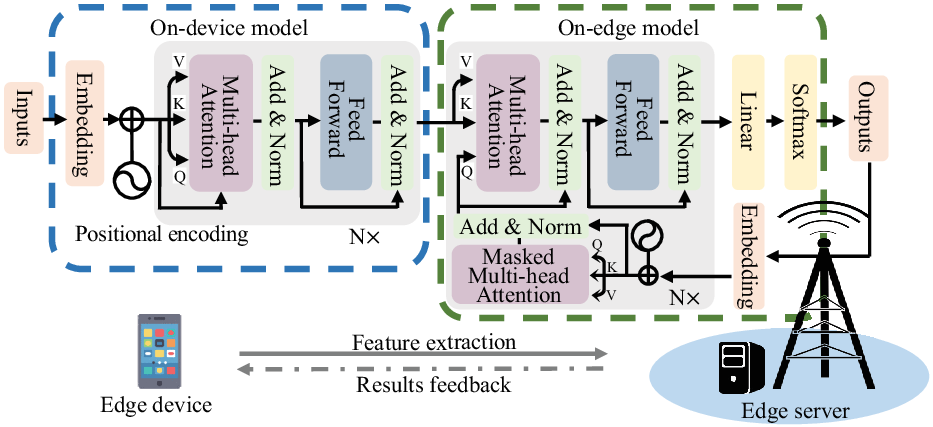}
		\vspace{-3pt}
	\caption{\label{model}Illustration of the considered LAIM co-inference system.}
\vspace{-3pt}
\end{figure}

\section{System Model}
We consider an LAIM co-inference system with one edge server and one device as shown in Fig. \ref{model}, where the edge device and server cooperatively perform LAIM inference based on pre-trained LAIMs. This setup corresponds to many real-world application scenarios, such as users submitting multi-modal prompts to access LAIM services at the network edge. 

Specifically, the pre-trained LAIM is partitioned into two segments (i.e., on-device and on-server models), each composed of several functional modules. Let the pre-trained LAIM be represented by a parameter vector $\mv{w}_{\rm p} \in \mathbb{R}^{q_{\rm p}}$, where $q_{\rm p}$ denotes the total number of model parameters. Each parameter of $\mv{w}_{\rm p}$ is assumed to be quantized with $b$ bits for storage. We split $\mv{w}_{\rm p}$ into two parts, and denote the parameters of on-device and on-server sub-models as $\mv{w} \in \mathbb{R}^{q}$  and $\mv{v} \in \mathbb{R}^{s}$, respectively, such that $q + s = q_{\rm p}$ and $\mv{w}_{\rm p} = \{\mv{w}; \mv{v}\}$.

Then, we deploy them on the edge device and server, respectively. Under this setup, the  inference process involves two stages. The first stage is on-device inference, where the device computes intermediate embeddings via the on-device model from the input and transmits them to the server. The second stage is on-server inference, where the edge server completes the remaining inference task via the on-server model and returns the final results.

\subsection{On-device Inference and Embeddings Transmission}

We first introduce the on-device inference process performed at the edge device during forward propagation. For efficient inference, we further prune $\mv{w}$ to obtain a lightweight model $\hat{\mv{w}} \in \mathbb{R}^{\hat{q}}$. The pruning ratio is defined as $\rho = \hat{q} / q$\footnote{For ease of exposition, we define the pruning ratio as the \textit{remaining} parameter ratio.}. Denote ${\mv \phi}$ as the input data, then the output intermediate embeddings generated by the on-device model are given by $\mv{o}=f(\mv{\phi},\hat{\mv{w}})$, where $f(\cdot)$ denotes the computation process of the on-device model.

Next, the intermediate embeddings $\mv{o}$ are transmitted to the edge server for on-server processing. Let $\mv{x}$ represent the transmitted signal, and we assume each element $x_i$ of $\mv{x}$ has normalized power, i.e., $\mathbb{E}[|x_i|^2] = 1$, without loss of generality. Then, let  $p$ denote the transmit power, which satisfies the maximum power constraint, i.e.,
\begin{align}
0 \le p \le P_{\max}.
\end{align}

Furthermore, let $g$ denote the uplink channel power gain from the edge device to the  server, which is assumed to be unchanged during each transmission period. Then the achievable uplink communication rate (in bits-per-second (bps)) is
\begin{align}  \label{rate}
	r(p)=B\log_2 \left(1+\frac{g p}{ B N_0} \right),
\end{align}
where $B$ is the uplink bandwidth, and $N_0$ is the power spectrum density (PSD) of the additive white Gaussian noise (AWGN) at the receiver of the edge server.

\subsection{On-server Inference and Final Results Feedback}

We now introduce the on-server inference process to acquire the final LAIM inference results. The edge server also prunes its sub-model $\mv{v}$ to obtain a lightweight version $\hat{\mv{v}} \in \mathbb{R}^{\hat{s}}$, where the pruning ratio is defined as ${\tilde \rho} = \hat{s} / s$. It is worth noting that the overall model (with on-device and on-server parts) size (in bits) is $(\rho q+ {\tilde \rho} s )b$.

Upon receiving the intermediate embeddings $\mv{o}$ from the edge device, the server then completes the inference task and generates the final output as
\begin{align}
    \tilde{\mv{o}} = {\tilde f}(\mv{o}, \hat{\mv{v}}),
\end{align}
where ${\tilde f}(\cdot)$ denotes the processing function of the on-server model with pruned parameters $\hat{\mv{v}}$.
Finally, the edge server sends the final inference result $\tilde{\mv{o}}$ back to the edge device to complete the whole co-inference process.

\subsection{Inference Delay and Energy Consumption Analysis}

We analyze the end-to-end inference delay and energy consumption of the considered LAIM co-inference system.

1) On-device inference: Denote the original on-device computational workload (in FLOPs) as $N_{\rm FLOP}$. After model compression, we assume the remaining workload is proportional to the pruning ratio $\rho$, resulting in a compressed workload of $\rho N_{\rm FLOP}$. Denote $f$ and $c$ as the clock frequency and the number of FLOPs per cycle of the processor at the edge device, respectively. Then, the delay for on-device inference is
\begin{align}
	t^{\rm I}(\rho, f) = \frac{\rho N_{\rm FLOP}}{f c}.
\end{align}
Moreover, the corresponding energy consumption is
\begin{align}
	e^{\rm I}(\rho, f) = \eta \frac{\rho N_{\rm FLOP}}{c} \phi f^2,
\end{align}
where $\eta$ and $\phi$ represent the power usage effectiveness (PUE) and a chip-specific power coefficient of the edge device.

2) Embeddings uploading: 
Let $\theta$ denote the size of the intermediate embeddings $\mv o$ (in bits)\footnote{Note that in this work, we focus on analyzing the inference distortion induced by model pruning, and do not consider the quantization noise in the transmission of embeddings.}. According to the uplink communication rate defined in \eqref{rate}, the uploading duration of the embeddings $\mv o$ is
\begin{align}
	t^{\rm U}(p) = \frac{\theta}{r(p)},
\end{align}
and the energy consumption for uploading embeddings is 
\begin{align}
	e^{\rm U}(p) = \frac{p \theta}{r(p)}.
\end{align}

3) On-server inference:  
Let $\tilde{N}_{\rm FLOP}$ denote the total on-server computational workload before pruning. After compression with pruning ratio $\tilde{\rho}$, the workload becomes $\tilde{\rho} \tilde{N}_{\rm FLOP}$. Denote $\tilde{f}$ and $\tilde{c}$ as the clock frequency and FLOPs per cycle of the processor at the edge server, respectively. Then, the delay and energy consumption for on-server inference are given by
\begin{align}
	\tilde{t}^{\rm I}(\tilde{\rho}, \tilde{f}) &= \frac{\tilde{\rho} \tilde{N}_{\rm FLOP}}{\tilde{f} \tilde{c}}, \\
	\tilde{e}^{\rm I}(\tilde{\rho}, \tilde{f}) &= \tilde{\eta} \frac{\tilde{\rho} \tilde{N}_{\rm FLOP}}{\tilde{c}} \tilde{\phi} \tilde{f}^2,
\end{align}
where $\tilde{\eta}$ and $\tilde{\phi}$ denote the PUE and power coefficient of the edge server, respectively.
Generally, since the size of the final inference results is typically small and the downlink is less resource-constrained, we neglect the feedback delay and energy consumption from the server to the device.

4) Total delay and energy consumption: The total end-to-end inference delay including on-device embeddings computing, uploading, and final results obtaining is 
\begin{align}
T(\rho,f, p, \tilde \rho, \tilde f)= t^{\rm I}(\rho, f)+ t^{\rm {U}} (p) + {\tilde t}^{\rm I}(\tilde \rho, \tilde f),
\end{align}
while the total energy consumption is 
\begin{align}
E(\rho,f, p, \tilde \rho, \tilde f)= e^{\rm I}(\rho, f)+ e^{\rm {U}} (p) + {\tilde e}^{\rm I}(\tilde \rho, \tilde f).
\end{align}

\section{Model Output Distortion Approximation}
Due to the highly non-linear nature of AI models, it is generally hard to directly characterize the output distortion introduced by model pruning. To address this challenge, we first approximate the output distortion of FC DNNs, which provides useful insights on how pruning-induced model parameter distortion relates to the output distortion. Then, we extend the analysis to more general AI models to further examine the feasibility of the approximation.

\subsection{Approximation of Model Output Distortion on FC DNNs}

To approximate the output distortion of an $L$-layer FC DNN under the co-inference paradigm, we first model the FC DNN explicitly. Specifically, the on-device model is given by 
\begin{align}
	f(\mv{\phi},\mv{W})={\mv W}^{(L_{\rm D})}\sigma \big({\mv W}^{(L_{\rm D}-1)}\sigma(\cdots  {\mv W}^{(1)}{\mv \phi})\big),
\end{align}
and the on-server model is given as
\begin{align}
	{\tilde f}(\mv{\phi},\mv{W},\mv{V})\!=\!{\mv V}^{(L_{\rm E})}\sigma \big({\mv V}^{(L_{\rm E}-1)}\sigma (\!\cdots\!  {\mv V}^{(1)}{f(\mv{\phi},\mv{W})})\big),
\end{align}
where $L_{\rm D}$ and ${L_{\rm E}}$ denote the total numbers of layers of on-device and on-edge models, ${\mv W}=\{{\mv W}^{(1)}, \cdots,{\mv W}^{(L_{\rm D})}\}$ and  ${\mv V}=\{{\mv V}^{(1)}, \cdots,{\mv V}^{(L_{\rm E})}\}$ denote the sets of model parameters, and $\sigma(\cdot)$ represents the activation function.

To facilitate the FC DNN output distortion analysis under the co-inference paradigm, we consider the following assumptions on the input data and activation function.

{\bf Assumption 1} (Normalized input data): The input data are normalized as $\|\mv{\phi}\|^{2} \le 1$ before being fed into DNNs.

{\bf Assumption 2} (Smoothness of the activation function): The activation function $\sigma(\cdot)$ is 1-Lipschitz, i.e.,
\begin{align}
	\|\sigma(\mv x)-\sigma(\mv y)\| \le \|{\mv x}-{\mv y}\|.
\end{align}
Furthermore, we assume $\sigma(0)=0$. Such assumptions correspond to various well-acknowledged activation functions such as tanh, ReLu, and LeakyReLu.

Based on these assumptions, we formulate the following lemma on the output of the $l$-th layer of the FC DNN.

\begin{lemma}\label{lemma1}
Denote the $l$-th layer of the FC DNN $\mv{W}$ as ${\mv W}^{(l)}$, and the first $l$ layers of $\mv{W}$ as $\mv{W}^{(1:l)}$, then we have 
\begin{align}
	\|f(\mv{\phi},\mv{W}^{(1:l)})\|\le \prod_{j=1}^{l} \|\mv{W}^{(j)}\|_{\rm F}.
\end{align}
\begin{IEEEproof}
From the Cauchy-Schwarz inequality, we have 
\begin{align}
\|f(\mv{\phi},\mv{W}^{(1:l)})\|\le & \|\mv{W}^{(l)}\|_{\rm F} \|\sigma\big(f(\mv{\phi},\mv{W}^{(1:l-1)})\big)\| \nonumber \\
&\le  \|\mv{W}^{(l)}\|_{\rm F} \|f(\mv{\phi},\mv{W}^{(1:l-1)})\|,
\end{align}
where the last inequality follows from Assumption 2. Then, we apply the same inequality to $\|f(\mv{\phi},\mv{W}^{(1:l-1)})\|$ repeatedly. With normalized input data (as in Assumption 1), the proof is thus concluded.
\end{IEEEproof}
\end{lemma}

Based on Lemma \ref{lemma1}, we obtain the following lemma on the output distortion of the $l$-th layer of the FC DNN.
\begin{lemma}\label{Lemma_lemma2}
The output distortion of the $l$-th layer of an FC DNN model $\mv W$ is upper bounded by 
\begin{align}\label{lemma2}
&\|f(\mv{\phi},\mv{W}^{(1:l)})-f(\mv{\phi},\hat{\mv{W}}^{(1:l)})\|\nonumber \\
& \le \|\mv{W}^{(l)}-\hat{\mv{W}}^{(l)}\|_{\rm F} \prod_{j=1}^{l-1} \|\mv{W}^{(j)}\|_{\rm F} \nonumber\\
&\qquad + \|{\mv{W}}^{(l)}\|_{\rm F} \|f(\mv{\phi},\mv{W}^{(1:l-1)})-f(\mv{\phi},\hat{\mv{W}}^{(1:l-1)})\|. 
\end{align}
\begin{IEEEproof}
	See Appendix  A.
\end{IEEEproof}
	\end{lemma}

We apply \eqref{lemma2} to $\|f(\mv{\phi},\mv{W}^{(1:l-1)})-f(\mv{\phi},\hat{\mv{W}}^{(1:l-1)})\|$ repeatedly, which, using some algebraic manipulations, leads to the following proposition for the output embedding distortion of the on-device model.

\begin{proposition}\label{proposition1}
An upper bound on the output embedding distortion of the on-device model is
\begin{align}
\|f(\mv{\phi},\mv{W})-f(\mv{\phi},\hat{\mv{W}})\| \le  \sum_{l=1}^{L_{\rm D}}  A^{(l)}\|\mv{W}^{(l)} - \hat{\mv{W}}^{(l)}\|_{\rm F},
\end{align}
where $A^{(l)}=\prod_{i=1,i \neq l}^{L_{\rm D}}\|{\mv{W}}^{(i)}\|_{\rm F}$.
\end{proposition}

Finally, by denoting the total number of on-device and on-server layers as $L=L_{\rm D}+L_{\rm E}$, the whole parameter set as $\bm{\mathit{\Omega}}=\{\mv{W},\mv{V}\}$, and the pruned model as $\hat {\bm{\mathit{\Omega}}}$, we have the following theorem about the overall output distortion of FC DNNs under the co-inference paradigm.

\begin{theorem} \label{theorem1}
The upper bound on the output distortion for FC DNN co-inference is
\begin{align} \label{theorem1-1}
	& \|f(\mv{\phi},\mv{W},\mv{V})-f(\mv{\phi},\hat{\mv{W}},\hat{\mv{V}})\| \nonumber \\
	& \le \sum_{l=1}^{L_{\rm E}}  \tilde{A}^{(l)} \|\mv{V}^{(l)} - \hat{\mv{V}}^{(l)}\|_{\rm F}
	 +\|f(\mv{x},\mv{W})-f(\mv{x},\hat{\mv{W}})\| \nonumber \\
	&= \sum_{l=1}^{L} \hat{A}^{(l)} \|\bm{\mathit{\Omega}}^{(l)} - \hat{\bm{\mathit{\Omega}}}^{(l)}\|_{\rm F},
\end{align}
where $\tilde{A}^{(l)}=\prod_{i=1,i \neq l}^{L_{\rm E}}\|{\mv{W}}^{(i)}\|_{\rm F}$ and $\hat{A}^{(l)}=\prod_{i=1,i \neq l}^{L}\|{\bm{\mathit{\Omega}}}^{(i)}\|_{\rm F}$.
\end{theorem}

\begin{remark}
Theorem \ref{theorem1} indicates that the distortion of the model output $\|f(\mv{\phi},\mv{W},\mv{V})-f(\mv{\phi},\hat{\mv{W}},\hat{\mv{V}})\|$ is upper bounded by the distortion
of the model parameters $ \|\bm{\mathit{\Omega}} - \hat{\bm{\mathit{\Omega}}}\|_{\rm F}^2$.  Theorem \ref{theorem1} bridges model-level pruning with task-level performance distortion. It provides a theoretical foundation on the design of a  distortion function to facilitate rate-distortion analysis in Section IV. Specifically, since $\hat{A}^{(l)}$ of \eqref{theorem1-1} is independent of the pruning operation, we exclude it from the design of the distortion function. Consequently, a natural form of distortion function arising from  \eqref{theorem1-1} is
\begin{align}\label{distortion}
	d(\bm{\mathit{\Omega}},\hat{\bm{\mathit{\Omega}}})=\|\bm{\mathit{\Omega}}-\hat{\bm{\mathit{\Omega}}}\|_{\rm F}.
\end{align}
In the following, we show that the distortion function in \eqref{distortion} is also applicable to other general AI models (such as LAIMs), where it effectively approximates the output distortion (measured by the $\ell_2$ norm) introduced by model pruning.
\end{remark}

\subsection{Approximation of Model Output Distortion on General AI Models}

For a general AI model, let $\bm{\mathit{\Omega}}^{\prime}$ and $\hat{\bm{\mathit{\Omega}}^{\prime}}$ denote the original and pruned models. The explicit expression of the model output $f(\mv{\phi},\bm{\mathit{\Omega}}^{\prime})$, as well as the distortion measurement $\|f(\mv{\phi},\bm{\mathit{\Omega}}^{\prime}) - f(\mv{\phi},\hat{\bm{\mathit{\Omega}}}^{\prime})\|$, is generally difficult to obtain due to the model's high non-linearity and structural complexity. 

To overcome this, we approximate the output distortion using the first-order Taylor expansion, i.e., 
\begin{align} \label{taylor}
f(\mv{\phi},\hat{\bm{\mathit{\Omega}}}^{\prime}) \approx f(\mv{\phi},\bm{\mathit{\Omega}}^{\prime}) 
+ \nabla_{\bm{\mathit{\Omega}}^{\prime}} f(\mv{\phi},\bm{\mathit{\Omega}}^{\prime}) 
: \left( \hat{\bm{\mathit{\Omega}}}^{\prime} - \bm{\mathit{\Omega}}^{\prime} \right),
\end{align}
where $:$ denotes the tensor contraction operation.

To further quantify this approximation, we make the following assumption on the gradient.

{\bf Assumption 3} (Gradient bound): The Frobenius norm of the gradient  is bounded by a positive constant $G$, i.e.,
\begin{align}
	\|\nabla_{\bm{\mathit{\Omega}}^{\prime}}f(\mv{\phi},\bm{\mathit{\Omega}}^{\prime})\|_{F} \le G.
\end{align}

\begin{figure*}[h]
    \centering 
	\subfigure[ResNet-152]{\includegraphics[width=.24\textwidth]{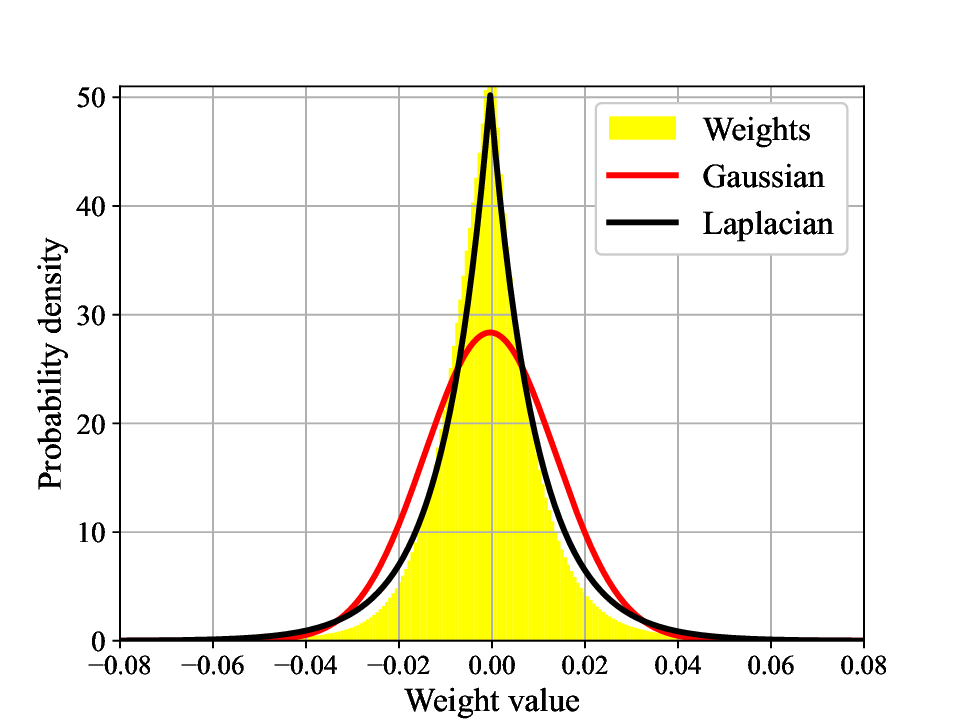}}
    \subfigure[BERT]{\includegraphics[width=.24\textwidth]{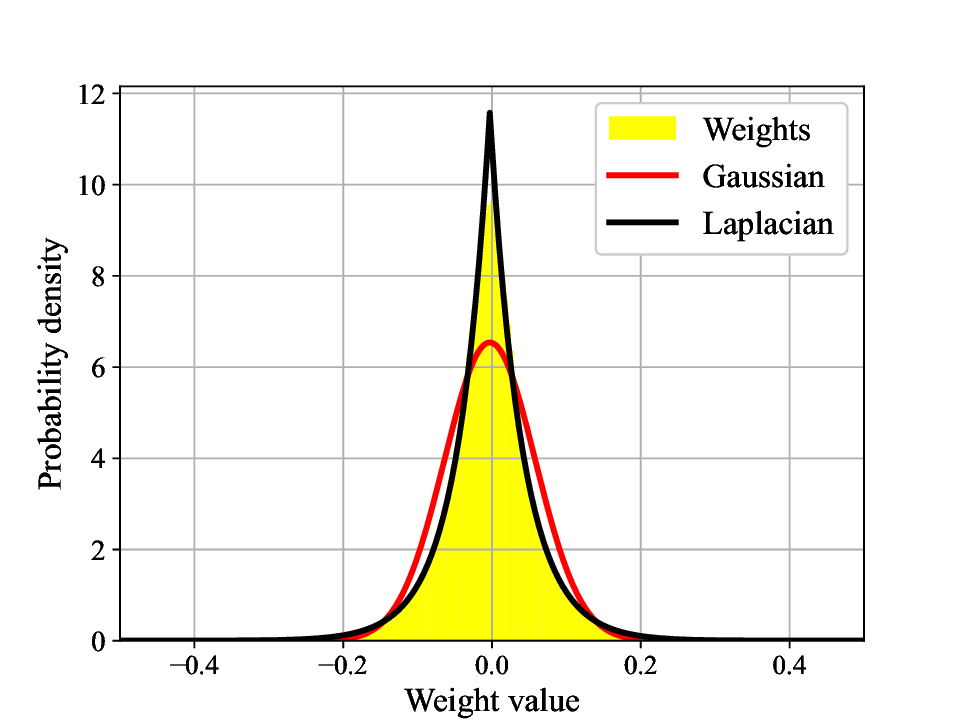}}
    \subfigure[BART]{\includegraphics[width=.24\textwidth]{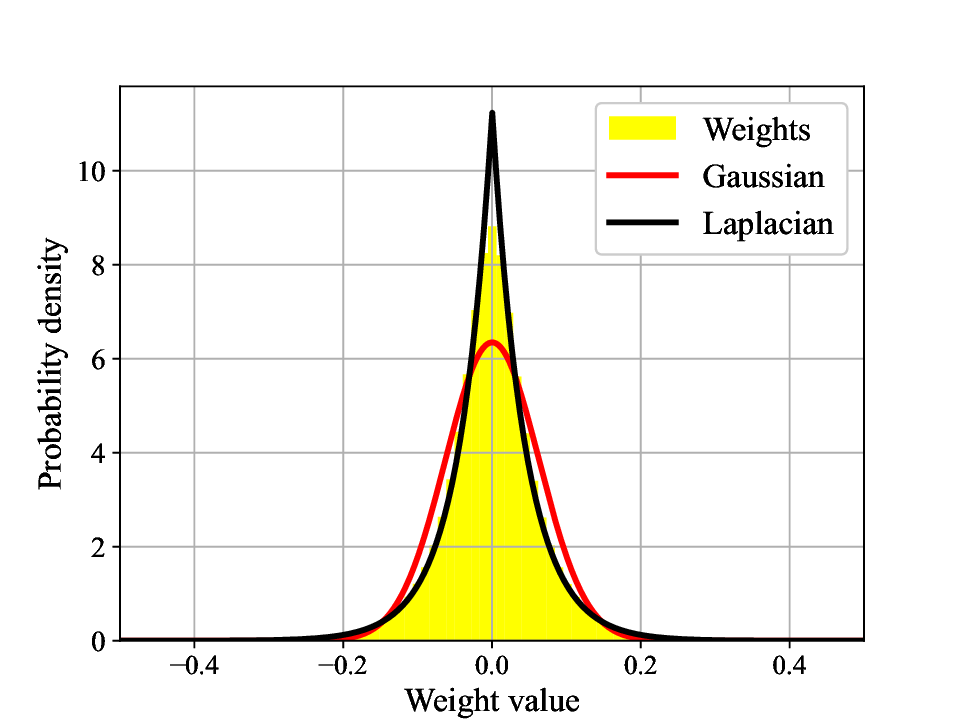}}
	\subfigure[GPT-3]{\includegraphics[width=.24\textwidth]{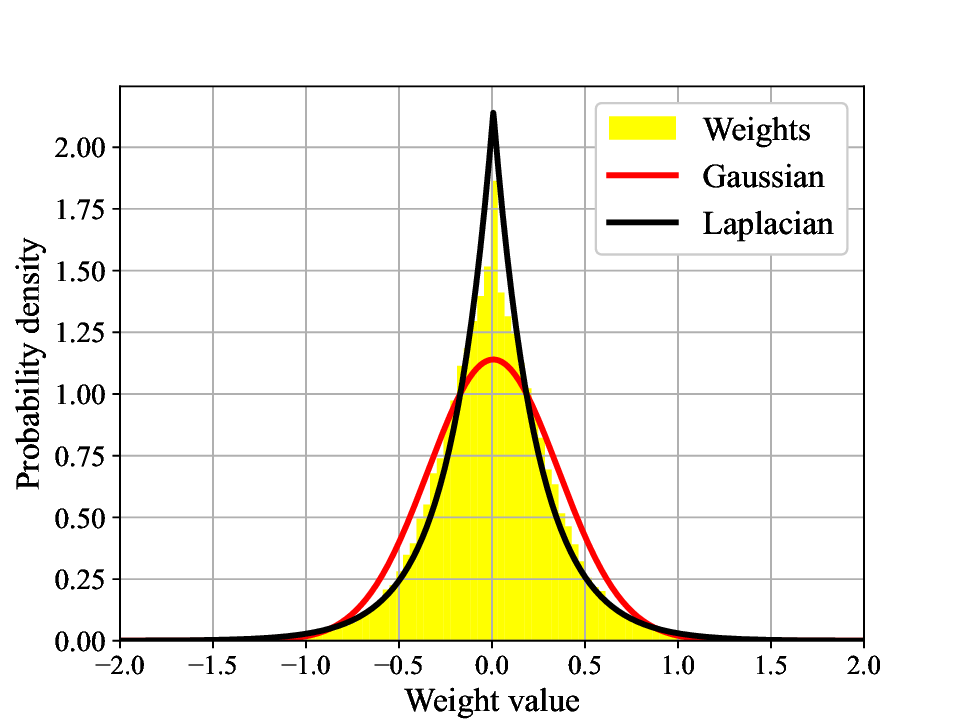}}
    \caption{Distribution of weights of various pre-trained models.}
    \label{fig:density_weight}
	\vspace{-7pt}
\end{figure*}

Under this assumption, the output distortion caused by pruning can be upper bounded as follows.

\begin{proposition} \label{proposition2.2}
	From \eqref{taylor} and Assumption 3, the upper bound on the output distortion of any general AI model $\bm{\mathit{\Omega}}^{\prime}$ is approximated as
	\begin{align}\label{distortion of non-linear}
	\|f(\mv{\phi},\hat{\bm{\mathit{\Omega}}}^{\prime}) - f(\mv{\phi},\bm{\mathit{\Omega}}^{\prime})\| &\approx \|\nabla_{\bm{\mathit{\Omega}}^{\prime}}f(\mv{\phi},\bm{\mathit{\Omega}}^{\prime}):({\bm{\mathit{\Omega}}^{\prime}}-\hat{\bm{\mathit{\Omega}}}^{\prime})\| \nonumber \\
	& = \| \tilde \nabla_{\bm{\mathit{\Omega}}^{\prime}}f(\mv{\phi},\bm{\mathit{\Omega}}^{\prime})\cdot {\rm vec} ({\bm{\mathit{\Omega}}^{\prime}}-\hat{\bm{\mathit{\Omega}}}^{\prime})\| \nonumber \\
	& \le G\|{\bm{\mathit{\Omega}}^{\prime}}-\hat{\bm{\mathit{\Omega}}}^{\prime}\|_F,
	\end{align}
	where $\tilde \nabla_{\bm{\mathit{\Omega}}^{\prime}}f(\mv{\phi},\bm{\mathit{\Omega}}^{\prime})$ is the matrix form of $\nabla_{\bm{\mathit{\Omega}}^{\prime}}f(\mv{\phi},\bm{\mathit{\Omega}}^{\prime})$ and ${\rm vec}(\cdot)$ denotes the matrix vectorization operation.
\end{proposition}

\begin{remark}
Proposition~\ref{proposition2.2} demonstrates that, for general AI models including transformer-based LAIMs, the output distortion caused by model pruning can be approximately upper bounded by the parameter distortion. This validates the use of the distortion function in \eqref{distortion} for broader model classes. Moreover, it provides theoretical justification for applying $d(\cdot)$ in \eqref{distortion} as a distortion function in our subsequent rate-distortion analysis in section IV, which characterizes the relationship between pruning ratio and LAIM co-inference performance.
\end{remark}

\section{Rate-distortion Analysis for Model Pruning}
To analyze the relationship between the model pruning ratio and inference performance, we aim to derive the rate-distortion function for LAIM co-inference. This  characterization could serve as a theoretical guidance on the design of pruning-aware LAIM co-inference systems.
However, such analysis is quite challenging. On one hand, as discussed in Section III, the output distortion of LAIMs is analytically intractable. Inspired by the results in Propositions \ref{proposition1} and \ref{proposition2.2}, we instead analyze the rate-distortion function of its upper bound, i.e., the model parameter distortion, under the distortion function in \eqref{distortion} and the weight distribution in Assumption 4. On the other hand, deriving the exact form of the rate-distortion function is also highly non-trivial, especially for high-dimensional data, such as LAIM parameters. To this end, we derive a lower bound, which provides a tractable approximation that effectively captures the relationship between the pruning ratio and model parameter distortion, and thereby reflects the expected output distortion in co-inference systems.

\subsection{Preliminaries of Rate-distortion Analysis for Pruning}

Before proceeding, we introduce the following definitions and assumptions for rate-distortion analysis.

{\it \underline{Definition} 4.1} (Achievable rate-distortion pair) \cite{TBerger} : Consider a source sequence ${\mv W}^{n}=\{\mv{w}_1,\cdots,\mv{w}_n\}$, which is a sequence of independent and identically distributed (i.i.d.) real-valued $q$-dimensional random vectors according to distribution $P_{\boldsymbol{w}}$. If there exists a series of encoder and decoder $\big(e_n(\cdot), g_n(\cdot)\big)$, such that the alphabet of code-word has size $2^{nR}$, and the expected distortion constraint $D$ is met under certain distortion function $d(\cdot)$, i.e., ${\rm lim}_{n\rightarrow \infty} \mathbb{E}[d({\mv W}^n, g_n(e_n({\mv W}^n)))] \le D$, then a rate-distortion pair $(R,D)$ is achievable.

{\it \underline{Definition} 4.2} (Rate-distortion function) \cite{TBerger} : The rate-distortion function $R(D)$ equals to the infimum of rate $R$, such that rate-distortion pair $(R,D)$ is achievable.

{\it \underline{Definition} 4.3} (Distortion-rate function) \cite{TBerger} : The distortion-rate function $D(R)$ is the infimum of all distortions, such that $(R,D)$ is achievable for a given rate $R$.

\begin{remark}
The rate-distortion function $R(D)$ and distortion-rate function $D(R)$ are two equivalent approaches to depict the boundary of the rate-distortion region, which is the set of all achievable rate-distortion pairs. 
\end{remark}

Accordingly,  we define the rate-distortion function for model pruning as follows.

{\it \underline{Definition} 4.4} ($R(D)$ for model pruning): 
Assume that the set of model parameters $\mv w$ follows the distribution $P_{\boldsymbol{w}}$, based on the distortion function defined in \eqref{distortion}, the rate-distortion function for model pruning is defined as
\begin{align}\label{r-d-general}
R(D)=\min _{P_{\hat{\boldsymbol{w}} |\boldsymbol{w}}: \mathbb{E}_{\boldsymbol{w}, \hat{\boldsymbol{w}}}[\|\boldsymbol{w}- \hat{\boldsymbol{w}}\|] \leq D} I(\boldsymbol{w} ; \hat{\boldsymbol{w}}) .
\end{align}

Before deriving the rate-distortion function $R(D)$ for model pruning, we first introduce a source distribution that characterizes the statistical properties of the pre-trained model weights.

{\bf Assumption 4} (Parameter Distribution): 
The parameter vector ${\mv w} \in \mathbb{R}^{q}$ of the pre-trained model is assumed to follow a multivariate Laplacian distribution with mean vector $\bm{\mu}$ and scale parameter $\lambda$. The probability density function (PDF) is given by \cite{TEltoft2006}
\begin{align} \label{multiLap}
P_{\boldsymbol{w}}(\mv{w}) = \frac{1}{(2 \pi)^{q / 2}} \cdot \frac{2}{\lambda} \cdot 
\frac{K_{(q/2)-1}\left(\sqrt{\frac{2}{\lambda} y(\mv{w})}\right)}{\left( \sqrt{\frac{\lambda}{2} y(\mv{w})} \right)^{(q/2)-1}},
\end{align}
where $K_{\alpha}(\cdot)$ denotes the modified Bessel function of the second kind of order $\alpha$, $y(\mv{w}) = (\mv{w} - \bm{\mu})^{\rm T} \bm{\mathit{\Gamma}}^{-1} (\mv{w} - \bm{\mu})$, and $\bm{\mathit{\Gamma}}$ is the covariance matrix of the distribution.

\begin{remark}
The assumption that the model parameters follow a Laplacian-type distribution is empirically supported. As shown in Fig.~\ref{fig:density_weight}, we analyze the empirical distributions of weights from various pre-trained models. It is observed that the Laplacian distribution provides a better fit than the commonly assumed Gaussian distribution. Nonetheless, the multivariate Laplacian distribution offers a more general modeling, as it captures the intrinsic correlations among model parameters. Such an assumption is more theoretically desirable for rate-distortion analysis in the subsequent section.
\end{remark}

\subsection{A Lower Bound on Rate-distortion Function for Pruning}

In the following, we derive a lower bound on the rate-distortion function to analyze the relationship between the pruning ratio and model parameter distortion. First, we introduce the following lemma on a lower bound of the rate-distortion function.
\begin{lemma}
Define $\boldsymbol{z} ={\mv w}-\hat{{\mv w}}$ as the pruning error following from a distribution $P_{\boldsymbol{z}}$. The rate-distortion function $R(D)$ of model weights is lower bounded by
\begin{align} \label{lemma-RD}
R(D) \ge h({\mv w})- \Phi(D),
\end{align}
where 
\begin{align}
\Phi(D) \triangleq \sup _{P_{\boldsymbol{z}}: \mathbb{E}\left[\|{\boldsymbol z}\|\right] \leq D} h({\mv z}),
\end{align}
and $h(\boldsymbol{w})$ is the differential entropy of $\mv{w}$, i.e., 
\begin{align}
h(\mv w)=-\int_{{\mathbb R}_q} P_{\boldsymbol{w}}(\mv w) \log \big({P_{\boldsymbol{w}}(\mv w)}\big) d {\mv w} .
\end{align}
\end{lemma}

\begin{IEEEproof}
The lower bound of the mutual information is 
\begin{align}
I({\mv w},\hat{{\mv w}})&=h({\mv w})-h({\mv w}|\hat{{\mv w}}) \nonumber \\
& \ge h({\mv w})- h({\mv w}-\hat{{\mv w}}|\hat{{\mv w}}) \label{lemma r-d lower bound 1-1}\\
& \ge h({\mv w})- h({\mv w}-\hat{{\mv w}}) \label{lemma r-d lower bound 1-2}\\
& \ge h({\mv w})- \sup _{P_{\boldsymbol z}: \mathbb{E}\left[\|\boldsymbol z \|\right] \leq D} h(\boldsymbol z),
\end{align}
where \eqref{lemma r-d lower bound 1-1} follows from the fact that shifting the mean does not change differential entropy, and \eqref{lemma r-d lower bound 1-2} follows from condition entropy being no larger than the entropy.
\end{IEEEproof}

Next, we proceed with deriving $\Phi(D)$ in \eqref{lemma-RD}. Specifically, scaling ${\mv z}$ to $\tilde {\mv z}={\mv z}/D$ such that $\mathbb{E} [\|\tilde {\mv z}\| ]\le 1$, we have
\begin{align}\label{phiD}
\Phi(D) &= \sup _{P_{{\boldsymbol z}}: \mathbb{E}\left[\|{\boldsymbol z}\|\right] \leq D} h({\boldsymbol z}) \nonumber \\
&= \sup _{P_{\boldsymbol z}: \mathbb{E}\left[\|{\boldsymbol z}\| \right] \leq D} - \int_{{\mathbb R}_q} \frac{d F_{\tilde {\boldsymbol z}}({\boldsymbol z}/D)}{d {\boldsymbol z}} {\rm log} \left(\frac{d F_{\tilde {\boldsymbol z}}({\boldsymbol z}/D)}{d {\boldsymbol z}} \right)d {\boldsymbol z} \nonumber \\
& = \sup _{P_{\tilde {\boldsymbol z}}: \mathbb{E}\left[\|\tilde {\boldsymbol z}\| \right] \leq 1} - \int_{{\mathbb R}_q} P_{\tilde {\boldsymbol z}}(\tilde {\boldsymbol z}){\rm log} \left(\frac{1}{D^q} P_{\tilde {\boldsymbol z}} (\tilde {\boldsymbol z})\right) d {\tilde {\boldsymbol z}}\nonumber \\
&= \sup _{P_{\tilde {\boldsymbol z}}: \mathbb{E}\left[\|\tilde {\boldsymbol z}\|^2\right] \leq 1} h(\tilde {\boldsymbol z}) + {\rm log}(D^q) \nonumber \\
&= \Phi(1) + q{\rm log}(D),
\end{align}
where $F_{\tilde {\boldsymbol z}}(\cdot)$ is the cumulative distribution function of $\tilde {\boldsymbol z}$.

Then, we have the following lemma on $\Phi(1)$,
\begin{lemma}
The value of $\Phi(1)$ in \eqref{phiD} is given by
\begin{align} \label{phi1}
		\Phi(1)\!\!= \!\!\!\!\sup _{P_{ {\boldsymbol z}}: \mathbb{E}\left[\| {\boldsymbol z}\|^2\right] \leq 1} \!\!\!\!h( {\boldsymbol z}) \!=\! q{\log}\left(\frac{\sqrt{\pi}{\rm e}}{q}\right)\!+\!\log \left(\frac{2\Gamma(q)}{\Gamma(q/2)}\right).
	\end{align}
\end{lemma}

\begin{IEEEproof}
First, we define 
\begin{align}
Q_{\beta}({\boldsymbol z})=\frac{{\rm e}^{-\beta\|\boldsymbol z\|}}{B(\beta)}, (\beta >0),
\end{align}
where $B(\beta)$ is defined as
\begin{align}
B(\beta)=\int_{{\mathbb R}_q} {\rm e}^{-\beta \|\boldsymbol z \|}d{\boldsymbol z} &=\int_{{\mathbb R}_q} \int_{\|\boldsymbol z \|}^{\infty} \beta {\rm e}^{-\beta x} dx d\boldsymbol z  \nonumber\\
&= \int_0^{\infty} \beta {\rm e}^{-\beta x} \int _{\|\boldsymbol z \| \le x} d \boldsymbol z  dx \label{A-1}\\
&= \beta \frac{\pi ^{q/2}}{\Gamma(\frac{q}{2}+1)} \int_0^{\infty} x^{q}  {\rm e}^{-\beta x}  dx \label{A-2} \\
& = \frac{2\Gamma(q) \pi ^{q/2}}{\Gamma(q/2) \beta^q}, 
\end{align}
where \eqref{A-1} follows from Fubini's theorem, \eqref{A-2} follows from the properties of the Gamma function, where 
$\Gamma(x)=\int_0^{\infty} t^{x-1}{\rm e}^{-t}dt$. 
With the defined $Q_{\beta}(\mv z)$, we have 
\begin{align}\label{bound of phi1}
\Phi(1)&=\sup _{P_{ {\boldsymbol z}}: \mathbb{E}\left[\|{\boldsymbol z}\|\right] \leq 1} h({\boldsymbol z}) \nonumber \\
&=\sup _{P_{{\boldsymbol z}}: \mathbb{E}\left[\|{\boldsymbol z}\|\right] \leq 1} -D_{\rm KL}\big(P_{\boldsymbol z}({\boldsymbol z}) \| Q_{\beta}({\boldsymbol z})\big)+\mathbb{E}\left[\log \frac{1}{Q_{\beta}({\boldsymbol z})}\right] \nonumber \\
& \le \log{B(\beta)}+\beta/ \ln 2,
\end{align}
where the last inequality follows from the fact that the Kullback-Leibler (KL) divergence $D_{\rm KL}\big(P_z({\boldsymbol z}) \| Q_{\beta}({\boldsymbol z})\big)$ is non-negative. Moreover, the lower bound is achievable if we set $P_{\boldsymbol z}({\boldsymbol z})= Q_{\beta}({\boldsymbol z})$ such that  $D_{\rm KL}\big(P_z({\boldsymbol z}) \| Q_{\beta}({\boldsymbol z})\big)=0$.

Finally, we minimize  \eqref{bound of phi1} by setting its derivative  with respect to (w.r.t.) $\beta$ equal to $0$, i.e., 
\begin{align}
\frac{d \big(\log{B(\beta)}+\beta/ \ln 2 \big)}{d \beta}=0,
\end{align}
and then substitute the obtained optimal $\beta$ into \eqref{bound of phi1}. This completes the proof.
\end{IEEEproof}

Finally, by substituting \eqref{phiD} and \eqref{phi1} into \eqref{lemma-RD}, we have the following theorem.

\begin{theorem}\label{theorem2}
Consider a multivariate Laplacian source. Under  the distortion function in \eqref{distortion}, its rate-distortion function is lower bounded by
\begin{align} \label{R_D_in_theorem2}
	R(D) \ge h(\mv{w})+q{\log}\left(\frac{q}{\sqrt{\pi}{\rm e}D}\right)+\log \left(\frac{\Gamma(q/2)}{2\Gamma(q)}\right).
\end{align}
Accordingly, for the considered co-inference system, with the original parameter dimension as $q_{\rm p}=q + s$ and the pruned model rate (size) as $(\rho q+ {\tilde \rho} s )b$, we have the lower bound of its distortion-rate function as
\begin{align}
D(R) \!\!\ge\!\! \frac{q+s}{\sqrt{\pi}{\rm e}} 2^{-\frac{(\rho q+ {\tilde \rho} s )b-h(\boldsymbol{w})}{q+s}} \!\left(\!\frac{\Gamma(\frac{q+s}{2})}{2\Gamma(q+s)}\!\right)^{\! \frac{1}{q+s}} \! \!\!\!\!\!\triangleq \hat D(\rho,\tilde \rho).
\end{align}
\end{theorem}

{\it \underline{Remark} 4.3} : 
From Theorem \ref{theorem2}, we have the following
observations.
First,  $ D(R) $ and $ R(D) $ are inverse functions. Such duality relationship characterizes the inherent trade-off between model pruning and inference quality.
Second, Theorem~\ref{theorem2} explicitly shows the influence of key system parameters on the inference distortion, including the pruning ratios $\rho$ and $\tilde \rho$, the model quantization level $b$, and the differential entropy $h(\mv{w})$. In particular, smaller pruning ratios $\rho$ and $\tilde \rho$ lead to reduced distortion, but at the cost of increased delay and energy consumption for additional computation workloads. To this end, it is crucial to properly design pruning ratios, jointly with other communication and computation resources to balance the trade-offs among inference accuracy, latency, and energy consumption.
Finally, the bound shows that for the same pruning ratio, models with higher entropy (i.e., larger $h(\mv{w})$) experience a larger distortion after pruning. This implies that LAIMs with more information-rich parameters are more sensitive to pruning. Thus, task-oriented model selection is also beneficial for inference quality in resource-limited edge environments.

{\it \underline{Remark} 4.4 }(Special-case on parallel Laplacian sources): Here we further assume that the weights follow a parallel (independent but not necessarily identical) Laplacian distribution. Specifically, each parameter $w_i \in \mv{w}$ follows a Laplacian distribution with zero-mean and scale factor  $\lambda_i$, i.e.,
\begin{align}
P_w(w_i)=\frac{\lambda_i}{2} {\rm e}^{-{\lambda_i|w_i|}}.
\end{align}
Based on this, we analyze the rate-distortion function in the following Proposition.

\begin{proposition}\label{parallel Lap}
Consider a parallel Laplacian source $\mv{w}$, where each parameter $w_i \!\in\! \mv{w}$ follows a Laplacian distribution with zero-mean and scale factor $\lambda_i$.  Under the distortion metric in \eqref{distortion}, the rate-distortion function is lower bounded by
\begin{align}
	R(D) \ge \sum_{i=1}^{q} \max \{ - {\log} ({\lambda_i} \mu), 0\},
\end{align}
where $\mu$ is chosen as $\sum_{i: \mu < \frac{1}{\lambda_i}} \frac{\mu}{\sqrt{q}}+\sum_{i: \mu \ge \frac{1}{\lambda_i}} \frac{1}{\lambda_i \sqrt{q}} = D$.
\end{proposition}
\begin{IEEEproof}
See Appendix B.
\end{IEEEproof}

\section{Joint Pruning Ratio, Communication, and Computation Design for LAIM Co-inference}

\subsection{Problem Formulation}
This section jointly designs the pruning ratio, communication, and computation  resource management for LAIM co-inference. 
Specifically, we aim to jointly optimize the pruning ratios $\rho$ and $\tilde{\rho}$, the computational frequencies $f$ and $\tilde{f}$, and the transmit power $p$, to minimize the lower bound of the LAIM co-inference distortion in Theorem \ref{theorem2}. The optimization is performed under the constraints of inference delay and energy consumption, as well as available system resource constraints (such as the maximum transmission power and computation clock frequency). Accordingly, the problem is formulated as
\begin{subequations}\label{P1}
	\begin{align}
	\text{(P1)}:\mathop {\min }\limits_{\{\rho,f,p, \tilde \rho,\tilde f\}} &  ~\hat D(\rho,\tilde \rho) \nonumber\\
	\mathrm{s.t.}~&
	T(\rho,f, p, \tilde \rho, \tilde f) \le T_0 \label{P1-b}\\
	~
	&E(\rho,f, p, \tilde \rho, \tilde f) \le E_0 \label{P1-c}\\
	~
	& 0 \le \rho \le 1 \label{P1-d}\\
	~
	& 0 \le  \tilde \rho \le 1 \label{P1-e}\\
	~
	& 0 \le f \le f^{\rm max} \label{P1-f}\\
	~
	& 0 \le \tilde f \le \tilde f^{\rm max} \label{P1-g}\\
	~
	& (1) \nonumber ,
	\end{align}
	\end{subequations}
where \eqref{P1-b}  and \eqref{P1-c} indicate that the total inference delay and energy consumption should not exceed the given thresholds $T_0$ and $E_0$. By properly adjusting $T_0$ and $E_0$, we can  balance the trade-offs among the inference quality, as well as the delay and energy consumption. Moreover, \eqref{P1-d} $\sim$ \eqref{P1-g} specify the domains of the optimization variables, where $f_k^{\rm max}$ and $f_m^{\rm max}$ denote the maximum clock frequencies of the processors on the edge device and server. 
However, in problem (P1), the constraints in \eqref{P1-b} and \eqref{P1-c} are  non-convex, due to the close coupling  of  variables $\rho$, $f$, $p$, $\tilde \rho$, and $\tilde f$. Therefore,  problem (P1) is highly non-convex and non-trivial to be optimally solved in general.

\subsection{Proposed Solution of Problem (P1)}
We propose an effective algorithm to solve problem (P1). Specifically, the objective of problem (P1) is convex w.r.t the pruning ratios $\rho$ and $\tilde \rho$. To deal with the non-convex constraints in \eqref{P1-b} and \eqref{P1-c}, we first introduce the auxiliary variables $\rho ^ \prime$ and $\tilde \rho ^ \prime$. Then problem (P1) is reformulated as 
\begin{subequations}\label{P2}
	\begin{align}
	\text{(P2)}:\mathop {\min }\limits_{\{\rho,f,p, \tilde \rho,\tilde f,\rho ^ \prime, \tilde \rho ^ \prime \}} &  ~\hat D(\rho,\tilde \rho) \nonumber\\
	\mathrm{s.t.}~&\frac{N_{\rm FLOP} } {\rho^{\prime} f c} +  \frac{\theta}{B\log_2 \left(1+\frac{g p}{ B N_0} \right)} \nonumber \\
	& \qquad + \frac{ N_{\rm FLOP} }{\tilde \rho^{\prime} {\tilde f} {\tilde c}}  \le T_0\label{P2-a}\\
	~
	&\eta \frac{N_{\rm FLOP}}{\rho^{\prime} c} \phi {f}^2 +  \frac{p \theta}{B\log_2 \left(1+\frac{g p}{ B N_0} \right)} \nonumber \\
	& \qquad + {\tilde \eta} \frac{ {\tilde N}_{\rm FLOP}}{ \tilde \rho^{\prime} {\tilde c}} {\tilde \phi} {\tilde f}^2 \le E_0 \label{P2-b}\\
	~
	& \rho \le \frac{1}{\rho^{\prime}} \label{P2-c}\\
	~
	& \tilde \rho \le \frac{1}{\tilde \rho^{\prime}} \label{P2-d} \\
	~
	& (1), (44\rm c) \sim (44\rm f) \nonumber.
	\end{align}
	\end{subequations}
It is worth noticing that the constraint \eqref{P2-a} is convex w.r.t. $\rho ^ \prime$, $f$, $p$, $\tilde \rho ^ \prime$, and $\tilde f$. However, problem (P2) is still  non-convex  considering the non-convex constraints \eqref{P2-b} $\sim$ \eqref{P2-d}. To address them, we use the successive convex approximation (SCA) technique by approximating the non-convex terms with the first-order Taylor expansion in an iterative manner. For each iteration $k \ge 1$, denote the local points as $p^{(k)}$, $\rho^{\prime{(k)}}$, and $\tilde \rho^{\prime{(k)}}$. Then, the second term of the left-hand-side of \eqref{P2-b} is upper bounded by 
\begin{align} \label{p approximation}
&\frac{p{\theta}}{B \log_2 \left(1+\frac{g p}{ {B N_0}} \right)}  \nonumber \\
&\le \frac{p^{(k)}{\theta}}{B \log_2 \left(1+\frac{g p^{(k)}}{ B N_0} \right)} +  u(p^{(k)})(p-p^{(k)}) \buildrel \Delta \over = \zeta^{(k)}(p),
\end{align}
where 
\begin{align} \label{u in p}
&u(p^{(k)})=\frac{\theta}{B \log_2 \left(1+\frac{g p^{(k)}}{ B N_0} \right)}- \nonumber \\
&\frac{p^{(k)}{\theta}g}{B \log_2 \left( 1+\frac{g p^{(k)}}{ B N_0} \right)^2 ( B N_0 +g p^{(k)}) \ln2}.
\end{align}
Also, for the non-convex constraints in \eqref{P2-c} and \eqref{P2-d}, we approximate them based on their first-order Taylor expansions,  
\begin{align} 
&\rho - \frac{1}{\rho^{\prime}} \le  \rho - \frac{1}{\rho^{\prime^{(k)}}}+\frac{1}{\rho^{\prime^{(k)2}}}(\rho^{\prime}-\rho^{\prime^{(k)}}) \le 0, \label{44c approximation} \\
&\tilde \rho - \frac{1}{\tilde \rho^{\prime}} \le \tilde \rho - \frac{1}{\tilde \rho^{\prime^{(k)}}}+\frac{1}{\tilde \rho^{\prime^{({k})2}}}(\tilde \rho^{\prime}-\tilde \rho^{\prime^{(k)}}) \le 0. \label{44d approximation}
\end{align}

Finally, for the $k$-th iteration, we obtain the approximate convex version of problem (P2) in problem (P3.$k$), via substituting $\zeta^{(k)}(p)$ in \eqref{p approximation} into the second term of the left hand side of \eqref{P2-b}, and replacing \eqref{P2-c} and \eqref{P2-d} as \eqref{44c approximation} and \eqref{44d approximation}. Problem (P3.$k$) can be solved by standard convex optimization tools efficiently, such as CVX.
\begin{subequations}\label{P2}
	\begin{align}
	\text{(P3.{\it k})}:\mathop {\min }\limits_{\{\rho,f,p, \atop \tilde \rho,\tilde f,\rho ^ \prime, \tilde \rho ^ \prime \}} &  ~\frac{q+s}{\sqrt{\pi}{\rm e}} 2^{-\frac{(\rho q+ {\tilde \rho} s )b-h(\boldsymbol{w})}{q+s}} \left(\frac{\Gamma(\frac{q+s}{2})}{2\Gamma(q+s)}\right)^{\frac{1}{q+s}}  \nonumber\\
	\mathrm{s.t.}~&\eta \frac{N_{\rm FLOP}}{\rho^{\prime} c} \phi {f}^2 +  \zeta^{(k)}(p)+ {\tilde \eta} \frac{ {\tilde N}_{\rm FLOP}}{ \tilde \rho^{\prime} {\tilde c}} {\tilde \phi} {\tilde f}^2 \le E_0 \label{P3-a}\\
	~
	&\rho - \frac{1}{\rho^{\prime^{(k)}}}+\frac{1}{\rho^{\prime^{(k)2}}}(\rho^{\prime}-\rho^{\prime^{(k)}}) \le 0 \label{P3-b}\\
	~
	& \tilde \rho - \frac{1}{\tilde \rho^{\prime^{(k)}}}+\frac{1}{\tilde \rho^{\prime^{({k})2}}}(\tilde \rho^{\prime}-\tilde \rho^{\prime^{(k)}}) \le 0 \label{P3-c}\\
	& (1), (45\rm a), (44\rm c) \sim (44\rm f) \nonumber.
	\end{align}
	\end{subequations}
By iteratively solving problem (P3.$k$), we obtain a series of solutions, leading to monotonically non-increasing objective values of problem (P1). This guarantees the convergence of the proposed SCA-based algorithm for problem (P1). The iterative process stops when the decrease of the objective value is smaller than a given threshold $\epsilon$. 

\section{Numerical Results}

\begin{figure*}[h]
    \centering 
	\subfigure[FCDNN-8]{\includegraphics[width=.24\textwidth]{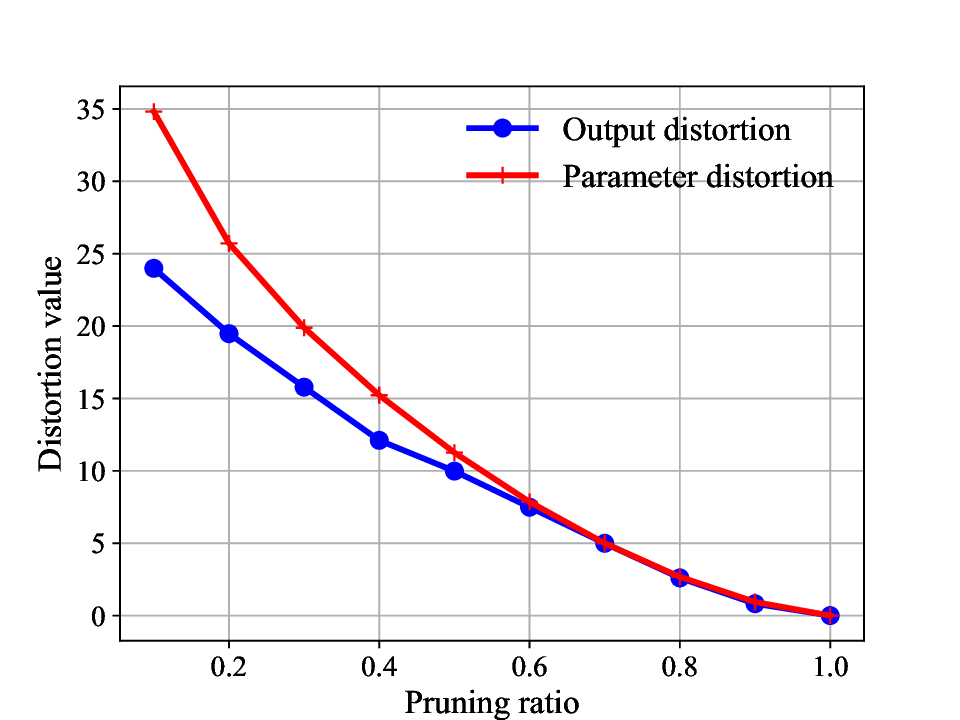}\label{fig:distortion_fcdnn8}}
        \subfigure[FCDNN-16]{\includegraphics[width=.24\textwidth]{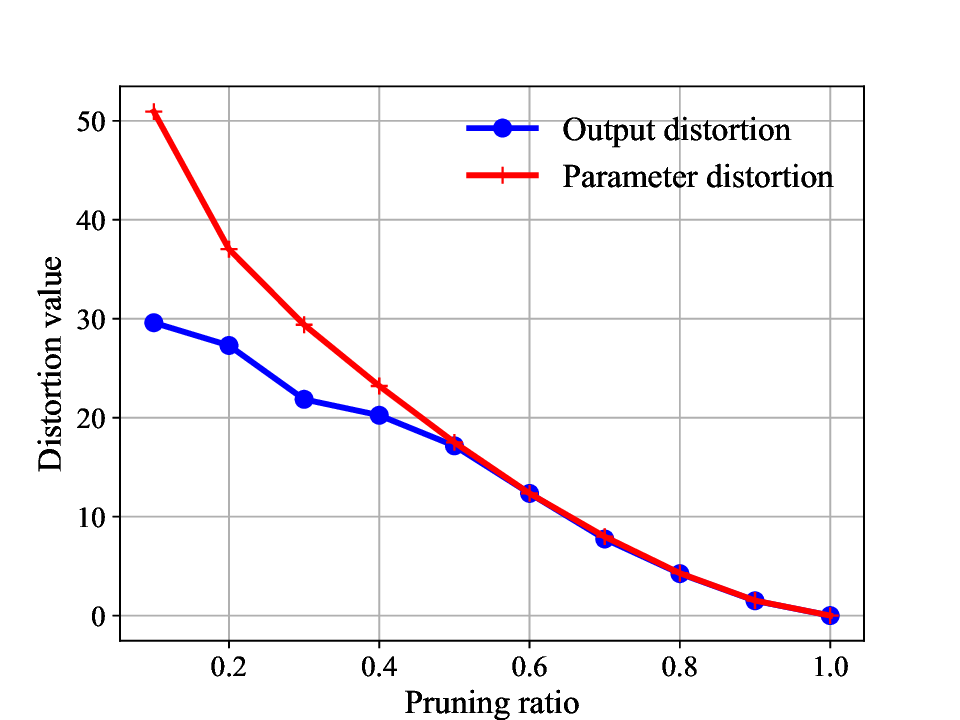}\label{fig:distortion_fcdnn16}}
        \subfigure[BERT]{\includegraphics[width=.24\textwidth]{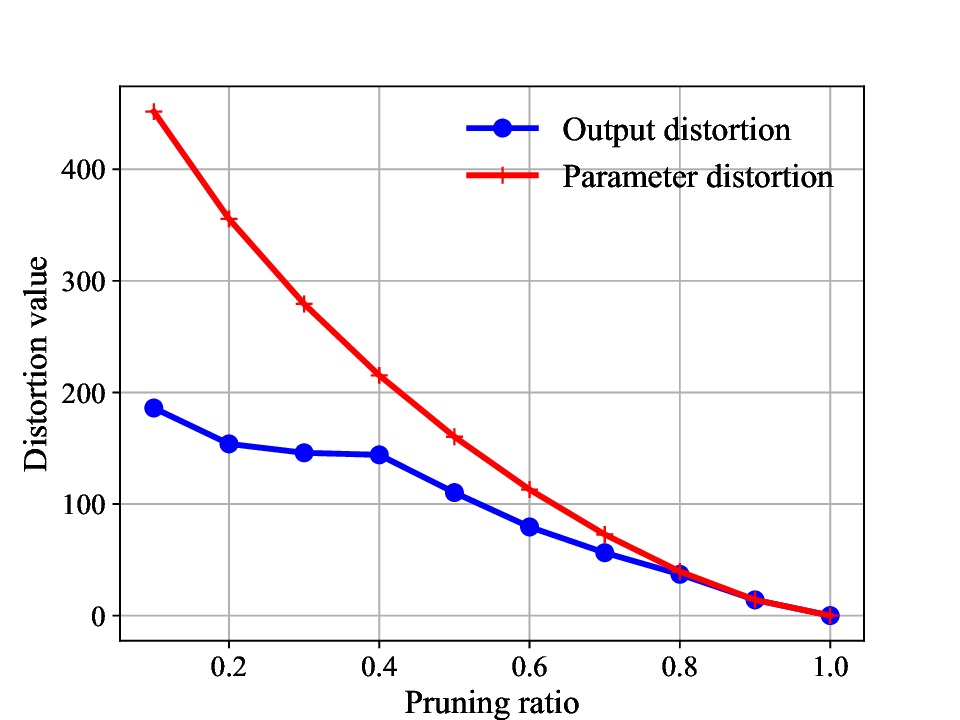}\label{fig:distortion_bert}}
		\subfigure[BART]{\includegraphics[width=.24\textwidth]{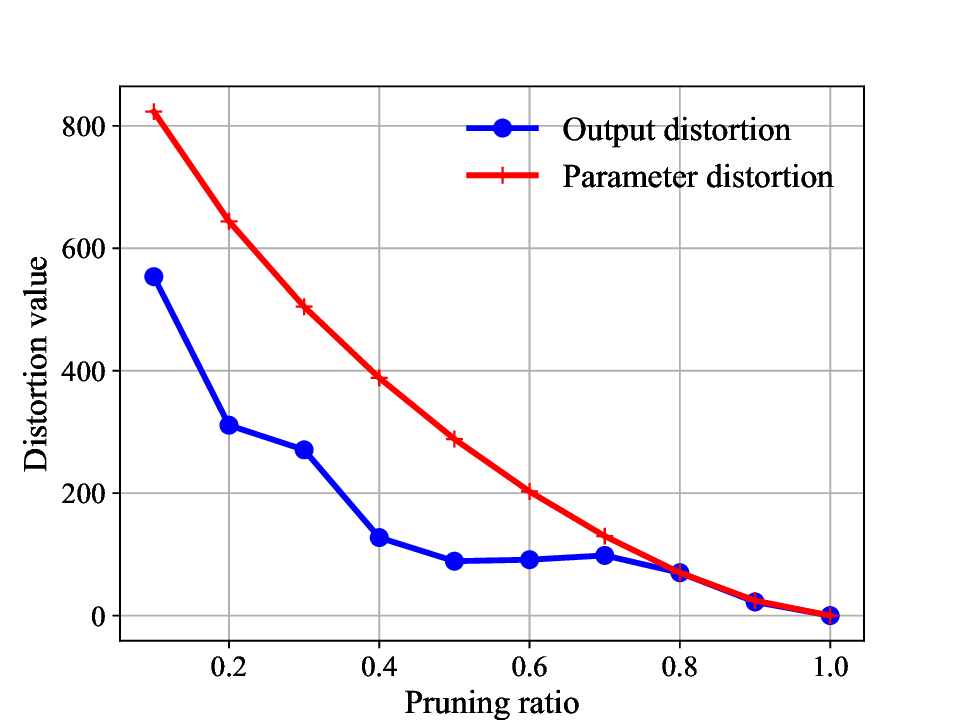}\label{fig:distortion_bart}}
    \caption{Distortions of model outputs and parameters w.r.t. pruning ratios.}\label{fig:distortion}
    \vspace{-2mm}
   \end{figure*}

We presents numerical results to validate the performance of the proposed joint pruning ratio, communication, and computation design for LAIM co-inference. All experiments were conducted with PyTorch on NVIDIA RTX 3090 GPUs. 


\subsection{Verification of Model Output Distortion Approximation}

We first verify the approximation for the model output  in Section III. Specifically, we start with verifying the approximated upper bound on FC DNNs with 8 and 16 hidden layers, denoted as FCDNN-8 and FCDNN-16, respectively. Then, we further validate the approximation on Google's BERT \cite{bert} and Facebook's BART \cite{bart}, which are two powerful widely-adopted LAIMs for natural language processing (NLP). Moreover, the adopted pruning strategy follows a widely-used magnitude-based criterion \cite{JLee}, which prioritizes the removal of weights with smaller magnitudes. We introduce the models in the following. 
\begin{itemize}
	\item {\bf FCDNN-8} adopts an auto-encoder architecture with 8 hidden layers trained on the MNIST dataset. The encoder has hidden sizes of $[64, 128, 256, 32]$, and the decoder has a symmetric structure. All layers employ the ReLU activation function.
	\item {\bf FCDNN-16}  is an auto-encoder with 16 hidden layers and ReLU activation functions. The encoder has 8 layers with sizes $[64, 128, 256, 512, 256,128, 64, 32]$, and the decoder is symmetric.  The deeper structure helps to evaluate the generality of the distortion approximation on more complex FC DNNs.
	\item {\bf Bidirectional Encoder Representations from Transformers (BERT)}  is a transformer-based model that leverages masked language modeling and next sentence prediction for pre-training. We adopt the base version (BERT-base) with 12 transformer encoder blocks, each with a hidden size of 768 and 12 attention heads. The model contains 109.48 million parameters and requires approximately 21.77 GFLOPs for a single inference pass.
	\item {\bf Bidirectional and Auto-Regressive Transformers (BART)}  combines a bidirectional encoder with an auto-regressive decoder, making it suitable for text generation and summarization tasks. It is trained as a denoising autoencoder, where the model reconstructs the original text from corrupted inputs. This allows BART to effectively learn both language understanding and generation capabilities. We use the base version (BART-base) with 6 transformer encoder blocks and decoder blocks, respectively, each with a hidden size of 1024. BART-base contains 139.42 million parameters and requires 6.44 GFLOPs per inference.
\end{itemize}

Fig.~\ref{fig:distortion} shows the gap between the model output distortion and the parameter distortion w.r.t. different pruning ratios across various model architectures. From Figs.~\ref{fig:distortion_fcdnn8} and~\ref{fig:distortion_fcdnn16}, even for highly non-linear models, such as FCDNN-8 and FCDNN-16, the parameter distortion serves as a reliable upper bound for the output distortion. In particular, when the pruning ratio is above 0.5, the gap becomes notably small, indicating that the parameter distortion can effectively approximate the output distortion.

For LAIMs like BERT and BART in Figs.~\ref{fig:distortion_bert} and~\ref{fig:distortion_bart}, the gaps between the two distortions increase. This is mainly due to the structural complexity of LAIMs, which makes the output distortion more difficult to approximate, thereby leading to relatively larger approximation errors.
Nevertheless, the distortion bound still remains tight when the pruning ratio is larger than 0.7. These observations confirm that the parameter distortion provides a practical and useful estimate of the output distortion for LAIMs, validating the generality of our analytical results on complex model architectures.

\begin{figure*}[h]
	\centering
	\subfigure[BART-Magnitude]{
	\centering
	\includegraphics[width=0.48\linewidth]{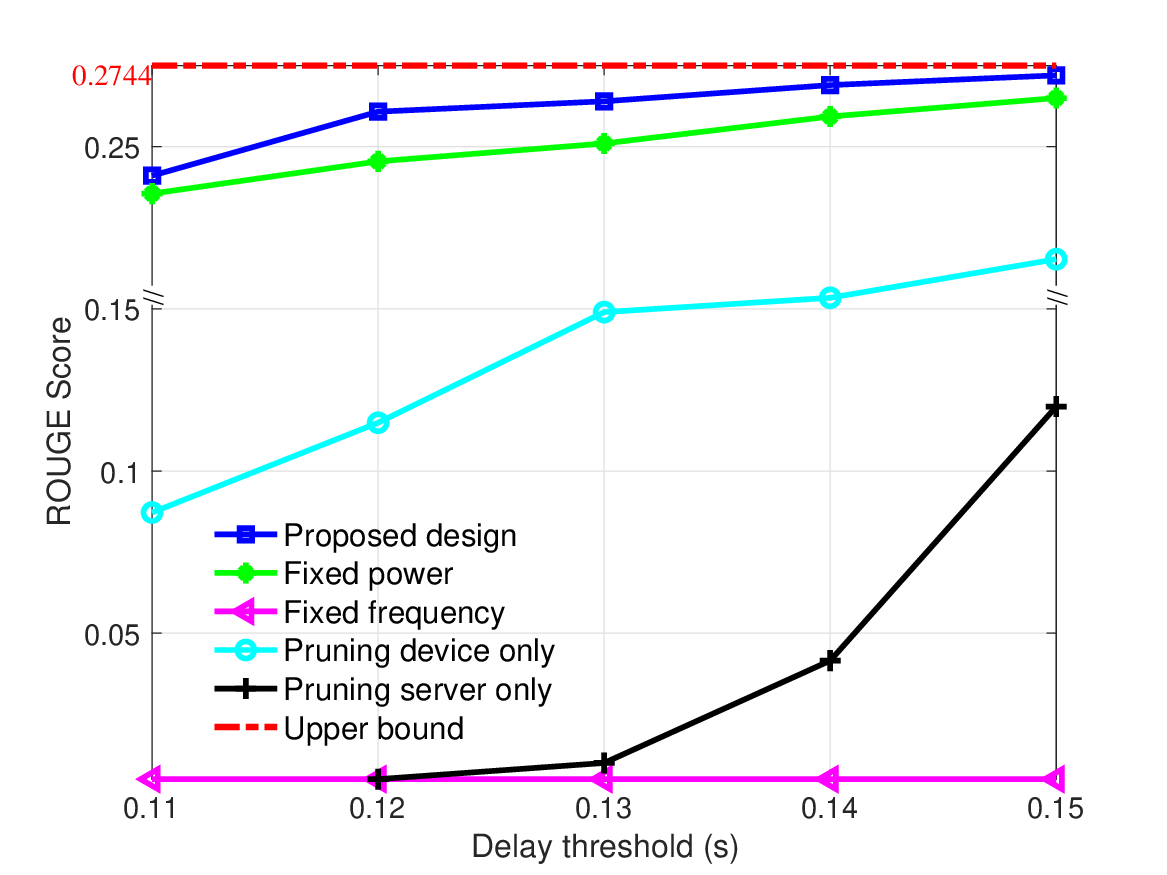} \label{delay-BART-Magnitude}
	}%
	\hfill
	\subfigure[BART-Random]{
	\centering
	\includegraphics[width=0.48\linewidth]{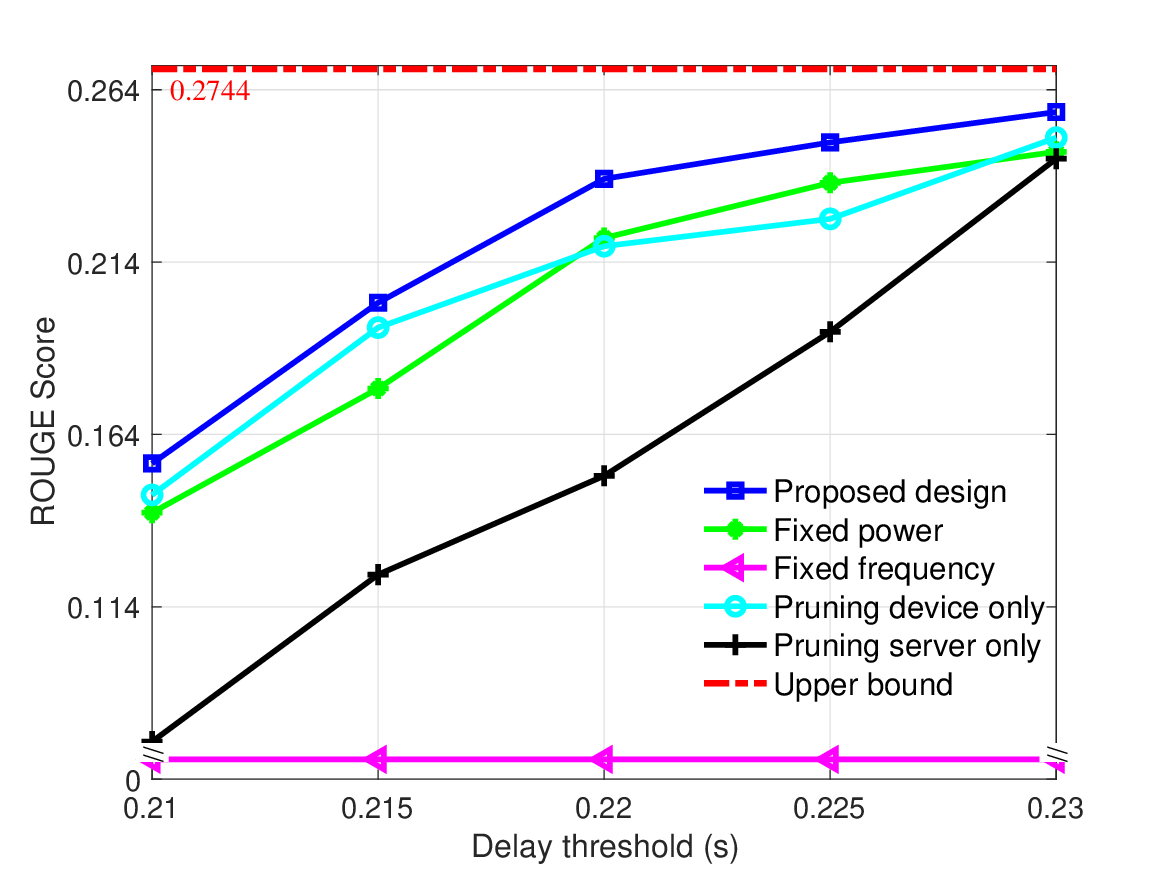}  \label{delay-BART-Random}
	} 

	\vspace{-5pt}

	\subfigure[BERT-Magnitude]{ 
		 \centering
	\includegraphics[width=0.48\linewidth]{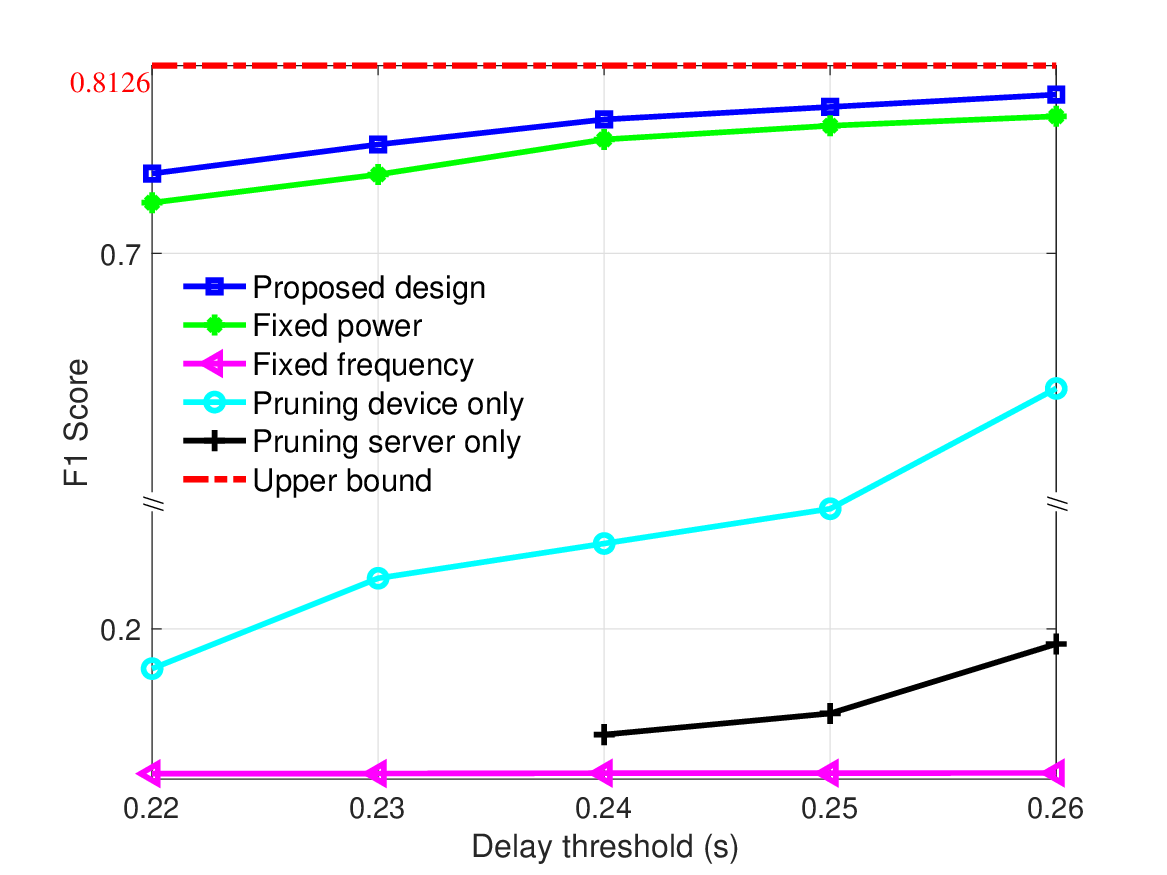}\label{delay-BERT-Magnitude}
	}%
	\hfill
	\subfigure[BERT-Random]{
	\centering 
	\includegraphics[width=0.48\linewidth]{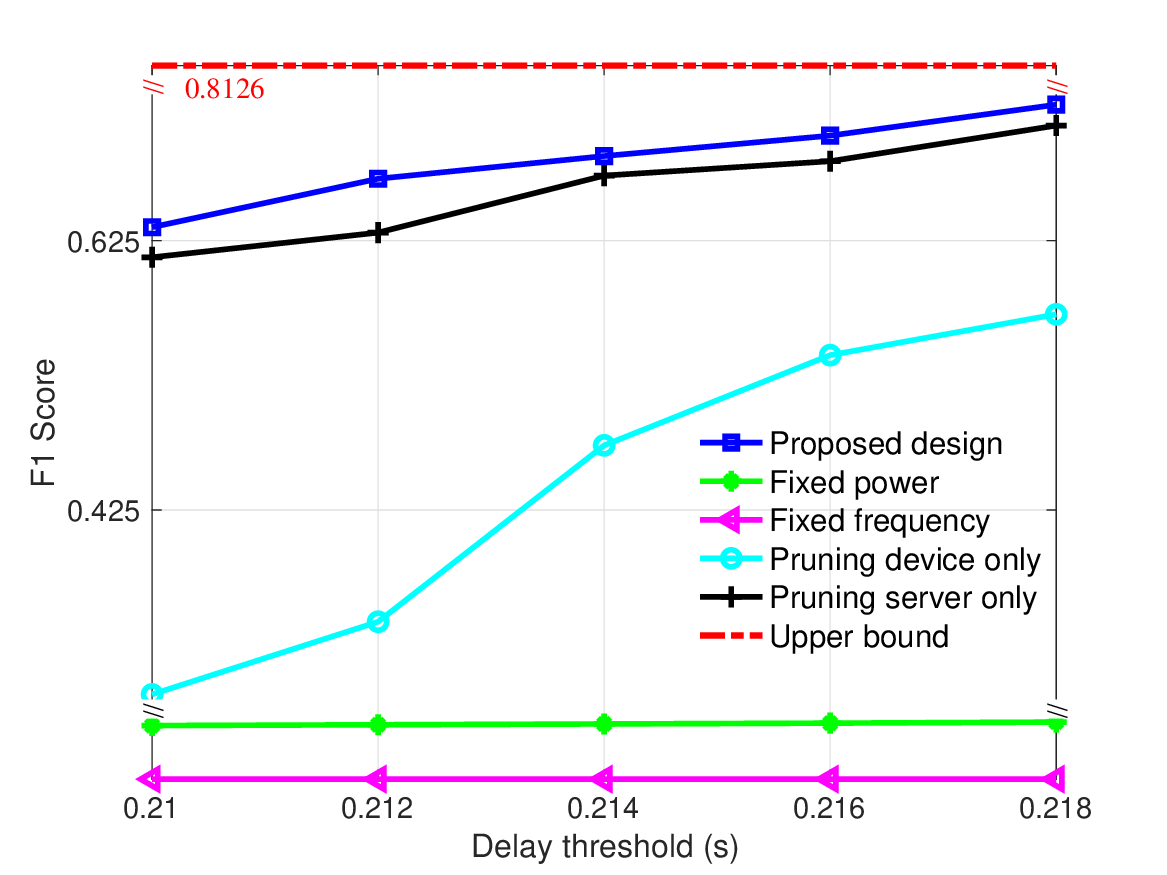}\label{delay-BERT-Random}
	} 
	\vspace{-5pt}
	\caption{System performance w.r.t. different delay thresholds. (a) Pruning BART with the magnitude paradigm under the inference energy consumption threshold $E_0 =0.0225 ~ {\rm J}$. (b) Pruning BART with the random paradigm under the inference energy consumption threshold $E_0 =0.0225 ~ {\rm J}$. (c) Pruning BERT with the magnitude paradigm under the inference energy consumption threshold $E_0 =0.0150 ~ {\rm J}$. (d) Pruning BERT with the random paradigm under the inference energy consumption threshold $E_0 =0.0011 ~ {\rm J}$.}
	\label{fig:delay}
	\vspace{-10pt}
\end{figure*}

\subsection{Performance of the Proposed Joint Pruning Ratio, Communication, and Computation Design}

\begin{figure*}[h]
	\centering
	\subfigure[BART-Magnitude]{
	\centering
	\includegraphics[width=0.48\linewidth]{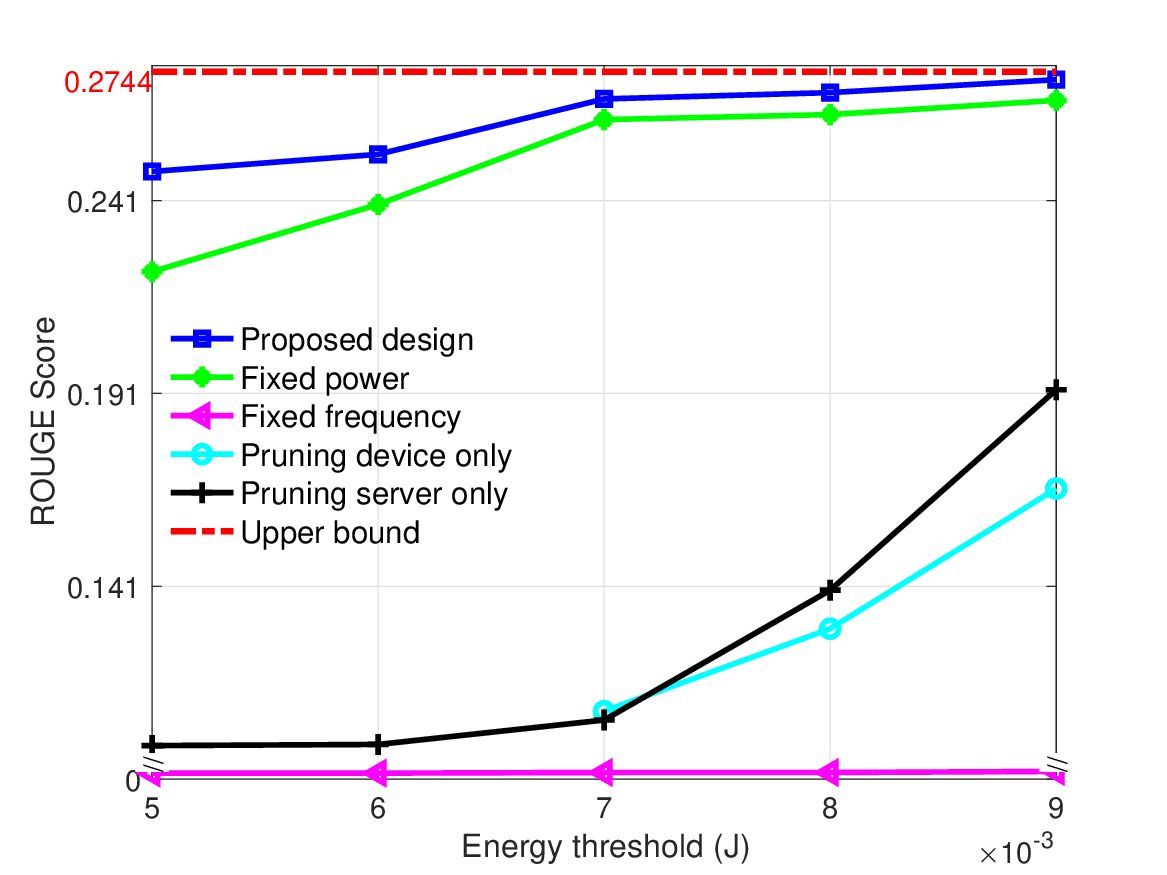} \label{energy-BART-Magnitude}
	}%
	\hfill
	\subfigure[BART-Random]{
	\centering
	\includegraphics[width=0.48\linewidth]{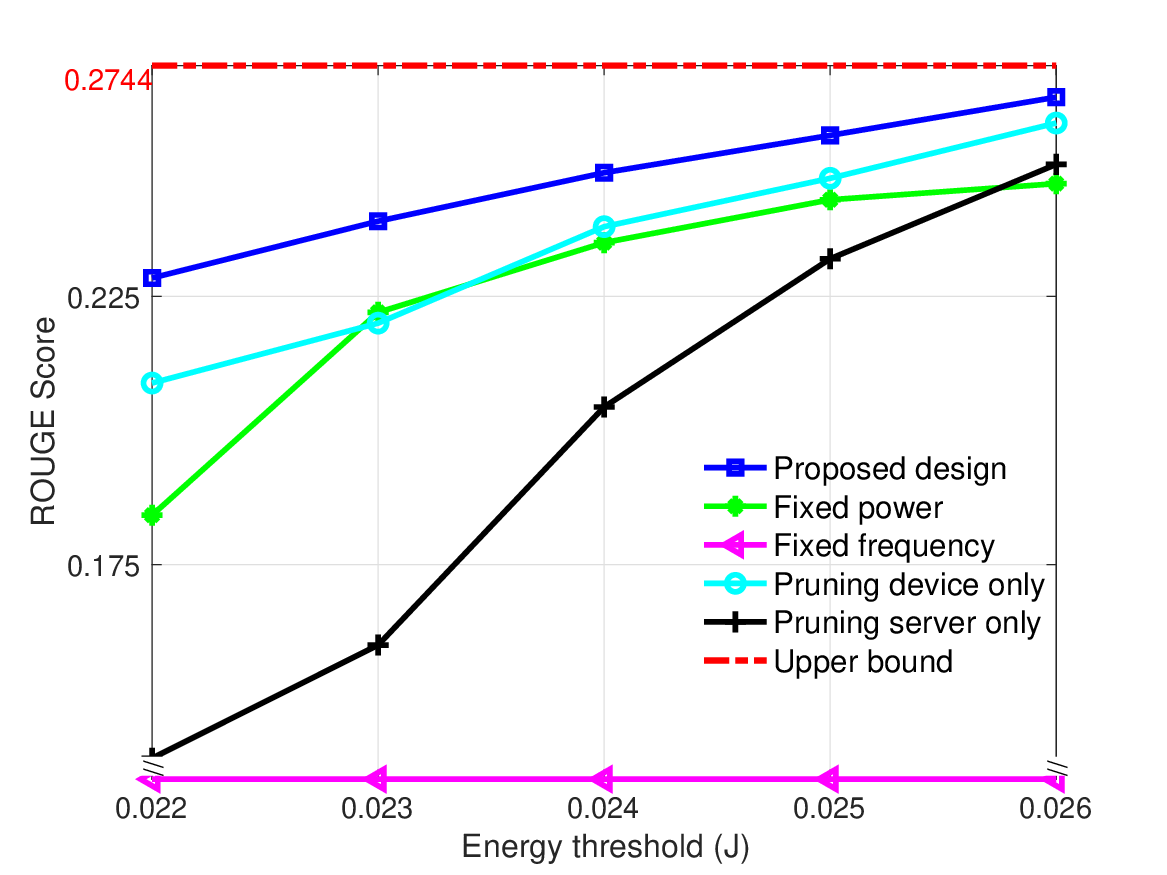}  \label{energy-BART-Random}
	} 

	\vspace{-5pt}

	\subfigure[BERT-Magnitude]{ 
		 \centering
	\includegraphics[width=0.48\linewidth]{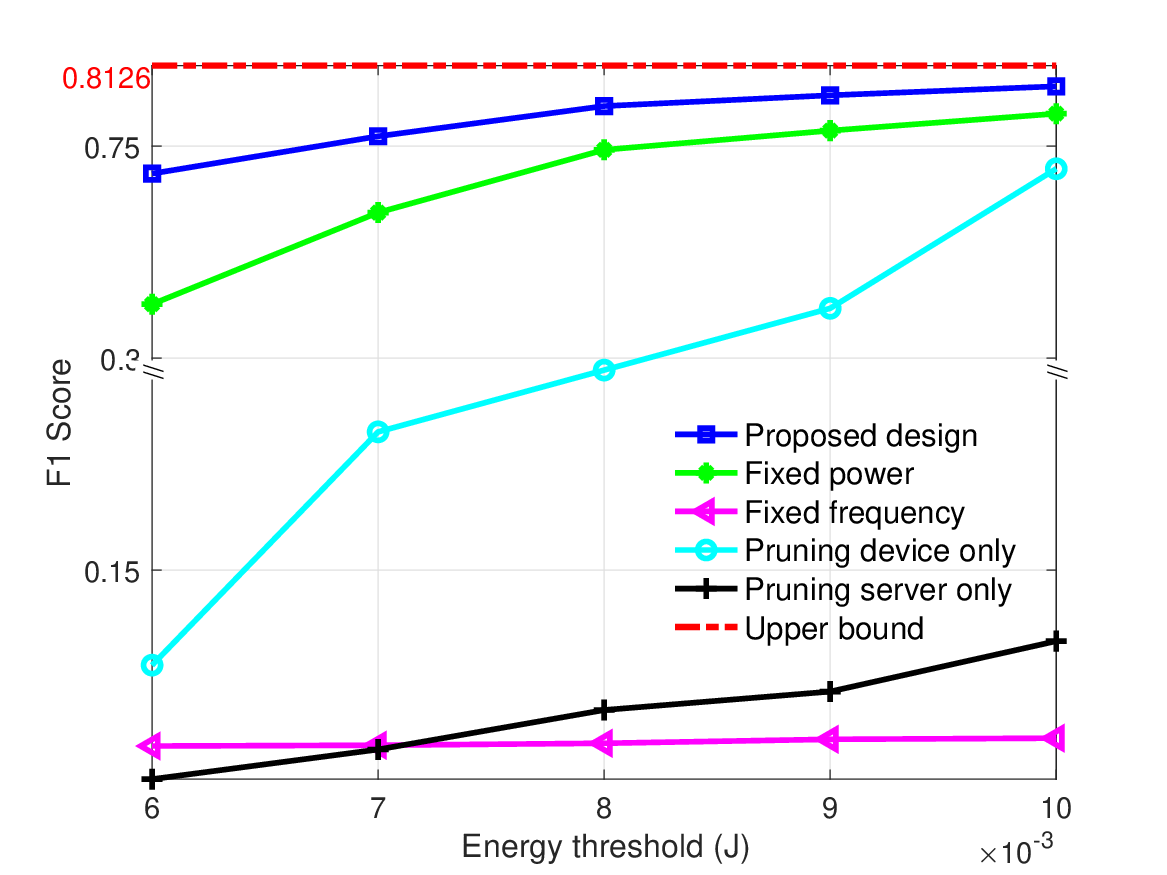}\label{energy-BERT-Magnitude}
	}%
	\hfill
	\subfigure[BERT-Random]{
	\centering 
	\includegraphics[width=0.48\linewidth]{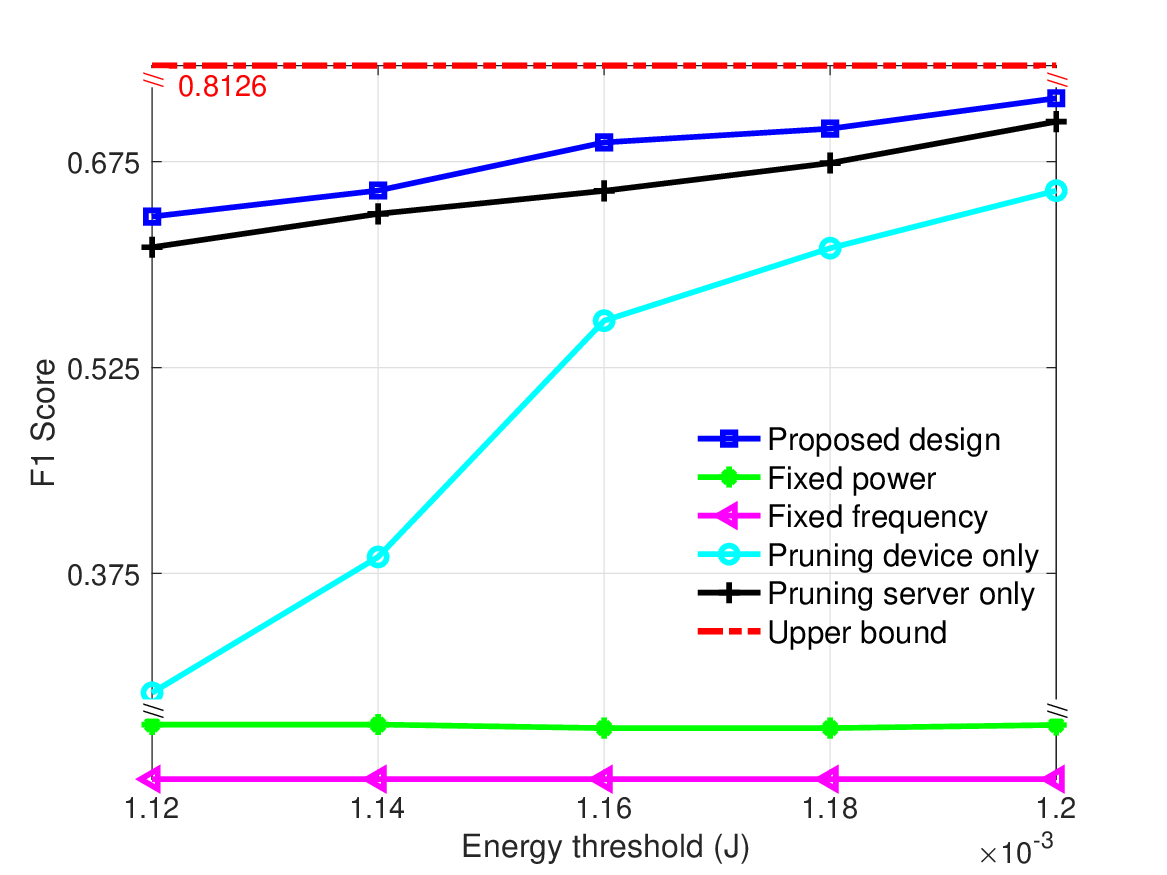}\label{energy-BERT-Random}
	} 
	\vspace{-5pt}
	\caption{System performance w.r.t. different energy consumption thresholds. (a) Pruning BART with the magnitude paradigm under the inference delay threshold $T_0 =0.22 ~ {\rm s}$. (b) Pruning BART with the random paradigm under the inference delay threshold $T_0 =0.22 ~ {\rm s}$. (c) Pruning BERT with the magnitude paradigm under the inference delay threshold $T_0 =0.35 ~ {\rm s}$. (d) Pruning BERT with the random paradigm under the inference delay threshold $T_0 =0.21 ~ {\rm s}$.}
	\label{fig:energy}
	\vspace{-5pt}
\end{figure*}

We evaluate the proposed joint pruning ratio, communication, and computation design for LAIM co-inference on BART and BERT models under different system delay and energy consumption requirements. Besides the magnitude-based pruning strategy, we additionally consider a random pruning scheme \cite{AHGadhikar}. Two widely used NLP datasets are employed for evaluation, i.e., CNN/DailyMail  \cite{cnn} and SQuAD \cite{squad}. The datasets and corresponding performance metrics are summarized as follows.
\begin{itemize}
	\item \textbf{CNN/DailyMail} is widely used for text summarization tasks. It contains online news articles from CNN and DailyMail. The dataset consists of 286,817 training samples, 13,368 validation samples, and 11,487 test samples. The evaluation metric is the ROUGE score, which measures the $n$-gram overlap between generated and reference summaries. Specifically, we consider ROUGE-1, ROUGE-2, and ROUGE-L, which evaluate unigram overlap, bigram overlap, and the longest common subsequence, respectively. We utilize the average ROUGE score as performance indicator.
	\item \textbf{Stanford Question Answering Dataset (SQuAD)} is a reading comprehension dataset constructed from Wikipedia articles. Each question is paired with a paragraph that contains the corresponding answer. The dataset includes 18,572 question-answer pairs for training and 2,322 samples for both development and testing. The performance metric is the $F_1$ score, which measures the harmonic mean of precision and recall scores, defined as
	\begin{align}
		F_1=2\times \frac{precision \times recall}{precision + recall}.
	\end{align}
\end{itemize}

We consider simulation parameters similar to those in \cite{ZLyu2025}. We set the maximum clock frequencies to $f^{\rm max} = 1~\rm{GHz}$ and $\tilde{f}^{\rm max} = 4~\rm{GHz}$, and the number of FLOPs per cycle to $c = 32$ and $\tilde{c} = 128$. We set the split point at the 4-th layer. For the wireless channel, we adopt a distance-dependent path-loss model as $K_0 (d/d_0)^{-\alpha}$, where $K_0 = -30~\rm{dB}$ is the reference path loss at distance $d_0 = 1~\rm{m}$, $\alpha = 2.8$ is the path-loss exponent, and the distance between the device and server is $d = 500~\rm{m}$. Moreover, we set the bandwidth to $B = 5~\rm{MHz}$, the maximum transmit power to $P_{\rm max} = 0.5~\rm{W}$, and the PSD of the AWGN to $N_0 = 2 \times 10^{-19}~\rm{W/Hz}$ (i.e., $-187~\rm{dB}$), resulting in a noise power of $-120~\rm{dBm}$. For energy modeling, we set the PUEs to $\eta = 1$ and $\tilde{\eta} = 2$, and the power coefficients to $\phi = 5 \times 10^{-29}~\rm{W/(cycle/s)^3}$ and $\tilde{\phi} = 1 \times 10^{-28}~\rm{W/(cycle/s)^3}$. The
following benchmark schemes are considered for performance
comparison.
\begin{itemize}
	\item {\bf Fixed power design:} We fix the transmit power as the maximum value, i.e., $p = {P}_{\rm max}$, and then solve problem (P1) for the remaining variables.
	\item {\bf Fixed clock frequency design:} We fix the clock frequencies to $f = f^{\rm max}$ and $\tilde f = \tilde f^{\rm max}$, and then optimize the remaining variables in problem (P1).
	\item {\bf Pruning on-device model only:} We only prune the on-device model with the full on-server model, i.e., solving problem (P1) by setting $\tilde \rho=1$.
	\item {\bf Pruning on-server model only:} We only prune the on-server model with the full on-device model, i.e., solving problem (P1) by setting $\rho=1$.
\end{itemize}

Fig.~\ref{fig:delay} shows the system performance w.r.t. varying delay thresholds for both BART and BERT models, evaluated with magnitude-based and random pruning strategies. First, it is observed that the overall performance improves as the delay thresholds increases. This is because higher delay thresholds allow for more complex computations and lower pruning intensity, thereby enhancing the inference performance.
Second, it is also observed that the proposed design outperforms all benchmark schemes, validating the effectiveness of joint optimization on pruning ratios, communication, and computation resources in LAIM co-inference. Moreover, as shown in Figs.~\ref{delay-BART-Magnitude} and~\ref{delay-BERT-Magnitude}, under stringent delay constraints, the pruning server only scheme becomes infeasible. This is due to the  relatively limited computational capacity of edge devices, which leads to excessive latency when performing full on-device model inference. When the delay threshold becomes the dominant system bottleneck, executing the entire  on-device model fails to meet real-time QoS requirements. In contrast, the proposed LAIM co-inference framework, with adaptive pruning and resource allocation, effectively accommodates tight latency requirements in edge scenarios.

Fig.~\ref{fig:energy} shows the system performance under different energy consumption thresholds. It is observed that the proposed design significantly outperforms the benchmark schemes. This demonstrates its efficiency in balancing the trade-offs between LAIM co-inference performance, system delay, and energy consumption. Furthermore, under stringent energy budgets, the  pruning device only scheme becomes infeasible (e.g., in Fig.~\ref{energy-BART-Magnitude}). This is because server-side processing generally consumes more energy due to some additional costs. When the energy becomes the dominant system bottleneck, executing the full on-server model may easily exceed the energy budget. In such cases, pruning only the on-device model is insufficient to meet the system's energy QoS requirements.

\subsection{Impact of Split Point}

\begin{figure}[h]
	\centering
	 \epsfxsize=1\linewidth
		\includegraphics[width=0.99\linewidth]{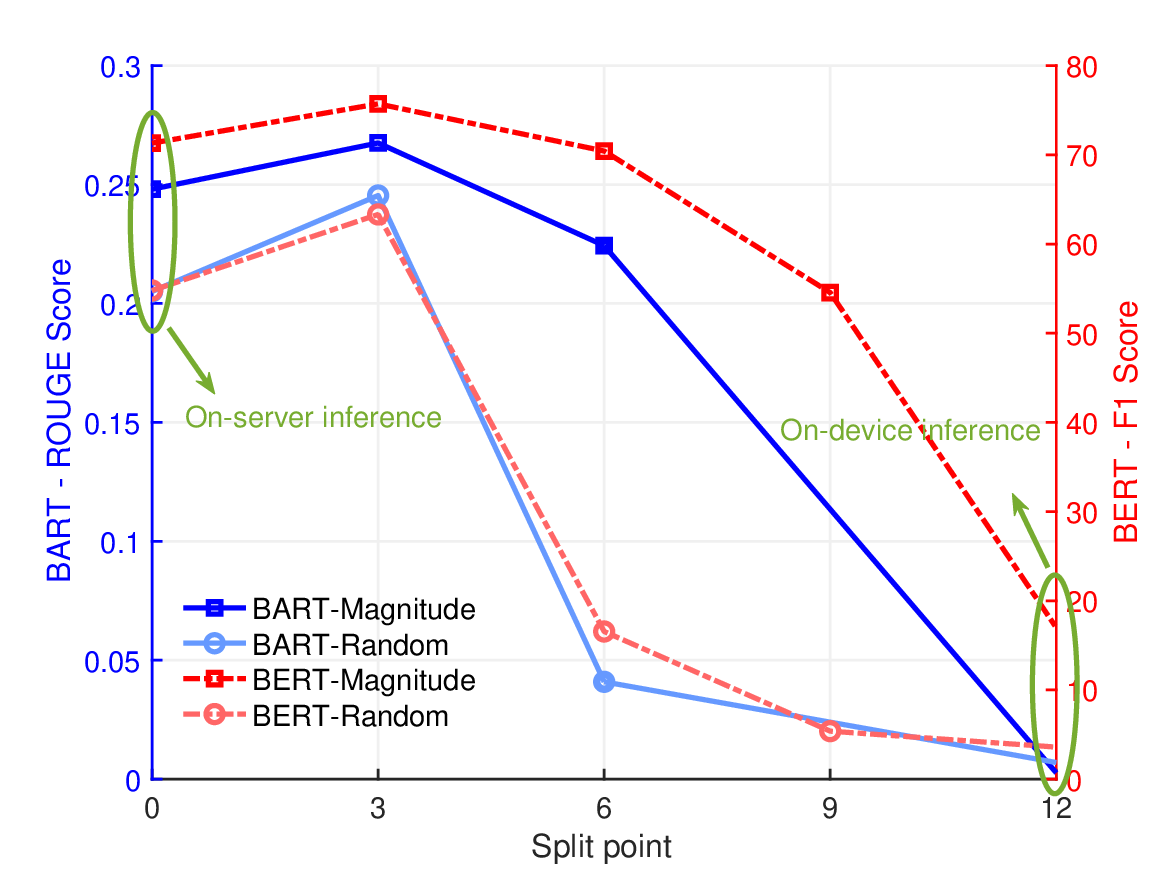}
		\vspace{-15pt}
	\caption{\label{difflayers}System performance w.r.t. different split points.}
	\vspace{-7pt}
\end{figure}

Fig.~\ref{difflayers} shows the system performance w.r.t. different split points in LAIM co-inference. Particularly, a split point of 0 corresponds to the on-server inference paradigm, where raw data samples are transmitted to the edge server for processing. In this case, only the server-side pruning ratio, clock frequency, and transmit power are optimized in problem (P1). A split point of 12 represents the on-device inference paradigm, where the entire inference process is conducted on the edge device locally, and only the device-side pruning ratio and clock frequency are optimized. Note that for BART-based co-inference, we split only the encoder, to avoid excessive back-and-forth transmission during token generation.

From Fig.~\ref{difflayers}, it is shown that the inference performance varies significantly across different split points, indicating the critical impact of split layer selection in co-inference system. Specifically, a small split point (i.e., few layers being deployed on the device) leads to underutilizing of the device computation resources. In contrast, when the split point is large, the limited processing capability of the device cannot efficiently handle the heavy computation workload, causing a dramatic performance drop. Therefore, it is important to properly set the split point for balancing the trade-offs among inference quality, system delay, and energy consumption. Finally, it is also shown that co-inference with a properly selected split point outperforms both on-device and on-server inference. This is because co-inference could better leverage the heterogeneous communication and computation resources of both the device and server, also with reduced communication overhead compared to transmitting raw data in on-server inference.

\vspace{-2pt}
\section{Conclusions}
We proposed a pruning-aware LAIM co-inference system, where a pre-trained LAIM is partitioned into on-device and on-server sub-models, and model pruning is applied to enable efficient inference under resource-constrained edge environments. For the proposed system, we first established that the output distortion of LAIMs is upper bounded by the parameter distortion, and then derived a lower bound on the parameter distortion via rate-distortion theory. Next, building upon the analytical result, we jointly optimized the pruning ratio, transmit power, and computation frequency to minimize the inference distortion bound, subject to constraints on system available resources, latency, and energy consumption. Then we developed an efficient algorithm to tackle the considered highly non-convex optimization problem. Finally, simulation results verified that our design significantly outperforms other benchmark schemes. 


\vspace{-5pt}

\begin{appendix}
\subsection{Proof of Lemma \ref{Lemma_lemma2}}
The output distortion of the $l$-th layer of the FC DNN is upper bounded by
\begin{align}
&\|f(\mv{\phi},\mv{W}^{(1:l)})-f(\mv{\phi},\hat{\mv{W}}^{(1:l)})\| \nonumber \\
& =\|\mv{W}^{(l)}\sigma\big(f(\mv{\phi},\mv{W}^{(1:l-1)})\big)-\hat{\mv{W}}^{(l)}\sigma\big(f(\mv{\phi},\hat{\mv{W}}^{(1:l-1)})\big)\| \nonumber \\
&= \|\mv{W}^{(l)}\sigma\big(f(\mv{\phi},\mv{W}^{(1:l-1)})\big)- \hat{\mv{W}}^{(l)}\sigma\big(f(\mv{W},\mv{W}^{(1:l-1)})\big) \nonumber \\
& \quad+ \hat{\mv{W}}^{(l)}\sigma\big(f(\mv{\phi},\mv{W}^{(1:l-1)})\big)- \hat{\mv{W}}^{(l)}\sigma\big(f(\mv{\phi},\hat{\mv{W}}^{(1:l-1)})\big)\| \nonumber \\
& \le \|\mv{W}^{(l)}\sigma\big(f(\mv{\phi},\mv{W}^{(1:l-1)})\big)- \hat{\mv{W}}^{(l)}\sigma\big(f(\mv{\phi},\mv{W}^{(1:l-1)})\big)\|\nonumber \\
& \quad + \|\hat{\mv{W}}^{(l)}\sigma\big(f(\mv{\phi},\mv{W}^{(1:l-1)})\big)- \hat{\mv{W}}^{(l)}\sigma\big(f(\mv{\phi},\hat{\mv{W}}^{(1:l-1)})\big)\| \label{lemma1-1}\\
& \le \|\mv{W}^{(l)}-\hat{\mv{W}}^{(l)}\|_{\rm F} \|f(\mv{\phi},\mv{W}^{(1:l-1)})\| \nonumber\\
&\qquad + \|\hat{\mv{W}}^{(l)}\|_{\rm F} \|f(\mv{\phi},\mv{W}^{(1:l-1)})-f(\mv{\phi},\hat{\mv{W}}^{(1:l-1)})\| \label{lemma1-2}\\
& \le \|\mv{W}^{(l)}-\hat{\mv{W}}^{(l)}\|_{\rm F} \prod_{j=1}^{l-1} \|\mv{W}^{(j)}\|_{\rm F} \nonumber\\
&\qquad + \|{\mv{W}}^{(l)}\|_{\rm F} \|f(\mv{\phi},\mv{W}^{(1:l-1)})-f(\mv{\phi},\hat{\mv{W}}^{(1:l-1)})\|,\label{lemma1-3}
\end{align}
where \eqref{lemma1-1} follows from the triangle inequality, \eqref{lemma1-2} follows from the Cauchy-Schwarz inequality and Assumption 2, \eqref{lemma1-3} follows from Lemma \ref{lemma1} and the fact that the Fobenius norm of pruned model parameters is always larger than it of the original model.

\subsection{Proof of Proposition \ref{parallel Lap}}
To start with, we introduce a lemma on the rate-distortion function on a Laplacian source under the distortion function $d(w, \hat w)= |w- \hat{w}|$.

\begin{lemma}\label{Lemma_appendix}
The rate-distortion function for a Laplacian source with zero-mean and scale factor of $\lambda$ under distortion function $d(w, \hat w)= |w- \hat{w}|$ is given by
\begin{align}
R(D)= \begin{cases}-\log (\lambda D), & 0 \leq D < \frac{1}{\lambda} \\ 0, & D \ge \frac{1}{\lambda}\end{cases}
\end{align}
\begin{IEEEproof}
	Please refer to \cite{TBerger} for more details.
\end{IEEEproof}
\end{lemma}

Next, we provide a lower bound on the rate-distortion function of a parallel Laplacian source {\mv w}. Specifically, the mutual information is lower bounded by
\begin{align}
I({\mv{w}, \hat{\mv{w}}})&=\sum_{i=1}^{q} h(w_i)- \sum_{i=1}^{q}  h(w_i|w_1,\cdots,w_{i-1},\hat{\mv w})\nonumber \\
&=\sum_{i=1}^{q} h(w_i)- \sum_{i=1}^{q}h(w_i| \hat{w}_i) \nonumber \\
&= \sum_{i=1}^{q} I (w_i| \hat{w}_i).
\end{align}

Then the rate-distortion function is lower bounded by 
\begin{align}
R(D) & = \mathop {\min }\limits_{{\mathbb E}\left[ \sqrt{\sum \limits\nolimits_{i=1}^{q} (w_i-{\hat w}_i)^2}\right]\le D} \sum \limits_{i=1}^{q} I (w_i| \hat{w}_i) \nonumber \\
& \ge \mathop {\min }\limits_{ \frac{\sum_{i=1}^{q} D_i}{\sqrt{q}}  \le D} \sum \limits_{i=1}^{q}  R(D_i)  \label{parallel_lap_Rd_1}\\
& = \mathop {\min }\limits_{ \sum_{i=1}^{q} D_i \le \sqrt{q}D} \sum \limits_{i=1}^{q}  {\max } \big\{-\log (\lambda_i D_i),0\big\}, \label{parallel_lap_Rd_2}
\end{align}
where $D_i=\mathbb {E} [|w_i-\hat{w}_i|]$, \eqref{parallel_lap_Rd_1} follows from the definition of the rate-distortion function and the fact that $\frac{\sum_i^q x_i}{q} \le \sqrt{\frac{\sum_i^q x_i^2}{q}}$, and \eqref{parallel_lap_Rd_2} follows from Lemma \ref{Lemma_appendix}. Then we construct Lagrangian function of the problem as
\begin{align}
{\cal L}(\{D_i\},\mu)= \sum \limits_{i=1}^{q} {\max } \big\{-\log (\lambda_i D_i),0\big\} + \mu D_i,
\end{align}
By setting the derivative w.r.t. $D_i$ equal to $0$, we obtain the final solution as 
\begin{align}
R(D) \ge \sum_{i=1}^{q} \max \{ - {\log} ({\lambda_i} \mu), 0\},
\end{align}
where $\mu$ is chosen as $\sum_{i: \mu < \frac{1}{\lambda_i}} \frac{\mu}{\sqrt{q}}+\sum_{i: \mu \ge \frac{1}{\lambda_i}} \frac{1}{\lambda_i \sqrt{q}} = D$.
Moreover, the corresponding distortion-rate function is 
\begin{align}
D(R) \ge \sum_{i=1}^{q} \min \{ \frac{\mu}{\sqrt{q}}, \frac{1}{\lambda_i \sqrt{q}}\},
\end{align}
where $\sum_{i: \mu < \frac{1}{\lambda_i}} - \log (\lambda_i \mu) = R$.

\end{appendix}

\end{document}